\documentclass[10pt,twocolumn,letterpaper]{article}

\usepackage{wacv}
\usepackage{times}
\usepackage{epsfig}
\usepackage{graphicx}
\usepackage{amsmath}
\usepackage{amssymb}

\usepackage{array}
\usepackage{booktabs}
\usepackage{subcaption}
\usepackage{url}

\wacvfinalcopy 

\def\wacvPaperID{316} 

\ifwacvfinal\fi
\begin{document}

\title{When Was That Made?}

\author{Sirion Vittayakorn \hspace{2cm} Alexander C. Berg \hspace{2cm} Tamara L. Berg\\
University of North Carolina at Chapel Hill\\
{\tt\small {sirionv, aberg, tlberg}@cs.unc.edu}
}

\maketitle
\ifwacvfinal\thispagestyle{empty}\fi

\begin{abstract} 
In this paper, we explore deep learning methods for estimating when
objects were made.  Automatic methods for this task could potentially
be useful for historians, collectors, or any individual
interested in estimating when their artifact was created. Direct
applications include large-scale data organization or
retrieval. Toward this goal, we utilize features from existing
deep networks and also fine-tune new networks for temporal estimation. In
addition, we create two new datasets of 67,771 dated clothing items
from Flickr and museum collections. Our method outperforms both a
color-based baseline and previous state
of the art methods for temporal estimation. We also provide several analyses
of what our networks have learned, and demonstrate applications to 
identifying temporal inspiration in fashion collections.
\end{abstract}

\vspace{-.4cm}
\section{Introduction}
\vspace{-.1cm}

Immense progress has been made on methods to categorize image and video
content. From recognizing objects~\cite{NIPS2012_4824,girshick2014rcnn}, to
scenes~\cite{places}, to activities~\cite{poseactionrcnn}, automatic
techniques are beginning to successfully answer the question of {\em what} is
shown in an image or video quite well. Recently, some computational methods
have begun to look at other aspects of the recognition problem, including
recognition of {\em where}~\cite{Hays:2008:im2gps}, or {\em when} an image was
captured~\cite{dating_colorim}.

In this paper, we explore the related problem of estimating when an object was
made from photographs of the object. Effective automatic methods for this task
could potentially be useful for historical dating at large scale by historians
and collectors, or in more everyday scenarios to identify the vintage of
objects. For example, online resellers like eBay and Etsy sell hundreds of
thousands of vintage furniture, clothing, and jewelry items. An automatic
method for dating man-made objects could be quite useful for organizing and
retrieving from these large disorganized collections.

The temporal periods of objects tend to be indicated by particular stylistic
elements, e.g., the shapes of headlights on cars, or the wide bell-bottoms of
pants from the 60s. As such, previous methods for temporal estimation have taken a
hand-crafted data mining approach to the problem, sifting through patches to discover
mid-level visual patterns that correlate with time
periods~\cite{lee_styleAware}. We take an alternative, purely discriminative,
approach to the problem using deep-learning based methods and achieve greater
accuracy than previous methods. Our first method trains discriminative models on top of
existing CNN features and yields the average mean absolute error (MAE) of $7.55\pm0.51$ years compared to the
previous state of the art of 8.56 years in~\cite{lee_styleAware} on their dataset of car images. 
Furthermore, by adapting and fine-tuning pre-tained networks to directly 
estimate the time, we improve performance to 3.97 years.
We also demonstrate the same procedure to train a deep network for temporal estimation 
of clothing on two new datasets collected for this paper,
consisting of 67K images of dated clothing items from the time period
of 1900 through 2009.

To our knowledge this is the first time deep networks have been used 
to estimate the time period of objects. As time has a natural
ordering, there are additional opportunities for introspection into
the structure learned by the deep network that are not available for
arbitrary classification tasks. To understand what the deep
networks are learning, we perform several analyses. First, we
demonstrate that fine-tuning the pre-trained network for
 temporal estimation of objects causes some nodes to become
significantly more selective to narrow time periods (we quantify this
over slices of the network using entropy histograms). 
We also perform analyses to compare the structures that the deep networks 
have learned to the previous mid-level discovery approach~\cite{lee_styleAware}.
Finally, we demonstrate the applicability of our models to identifying temporal 
influence in fashion collections.

In summary, our contributions are: 1) Deep learning approaches to estimate when an object was made, 2) Two new datasets of 67,771 dated photographs of clothing items made between 1900 and 2009; one collected from Flickr, and the other collected from museums, 3) Analyses of what the fine-tuned temporal estimation networks have learned, and comparison to the mid-level patterns learned by the previous state of the art approach~\cite{lee_styleAware}, and d) Application of temporal estimation to analyze the influence of vintage styles on fashion.

The rest of our paper is organized as follows. First, we review related works
(Sec~\ref{sec:related}). Then we describe three dated object datasets, one
existing, and two novel datasets (Sec~\ref{sec:datasets}). Next, we
describe our deep learning based approaches to estimate when objects were made
(Sec~\ref{sec:approaches}) and evaluate these models on the temporal estimation
task (Sec~\ref{sec:prediction}). Then, we present several analyses of what
our networks have learned and how our approach compares to a data mining approach to
temporal prediction (Sec~\ref{sec:analyses}). Finally, we apply our temporal estimation approach 
to explore the influence of vintage fashion on fashion show images (Sec~\ref{sec:app}).

\vspace{-.1cm}
\section{Related work}
\label{sec:related}
We review work in 3 areas related to our research: deep learning for visual
recognition, visual analyses of deep networks, and visual data mining.

\smallskip
\noindent{\bf Deep Learning:}
Convolutional neural networks (CNNs) have been one of the driving forces toward
improving performance of visual recognition algorithms. These deep learning
approaches, originally introduced as perceptrons in the 1950s~\cite{perceptron},
experienced a resurgence in popularity during the 1990s, and then in recent
years have almost taken over the recognition community after demonstrations of
remarkable image classification~\cite{NIPS2012_4824,simonyan15,he2015} and 
detection~\cite{googlenet} performance on benchmark tasks. We make use of both networks, from~\cite{NIPS2012_4824,simonyan15}, trained for classification of the 1.2 million labeled images in ImageNet~\cite{imagenet_cvpr09}. 
These networks demonstrate better performance than the best hand-crafted features on the
ImageNet Large Scale Visual Recognition Challenge (ILSVRC)~\cite{ILSVRC15}. 

The feature representations learned by these networks on ImageNet data have been shown to generalize well to other image classification
tasks~\cite{donahue2013decaf} as well as related tasks such as object detection~\cite{girshick2014rich,sermanet2013pedestrian}, pose estimation and action detection~\cite{poseactionrcnn}, or fine-grained category
detection~\cite{zhang14finegrained}. Moreover, in a somewhat related task to
ours, S. Karayev {\emph{et al.}}~\cite{karayev_bmvc14} show that using a pre-trained network~\cite{donahue2013decaf} as a generic feature extractor, produces a better classifier for photo and painting style than hand-crafted features. We are not aware of any prior work on modeling historical visual style using deep-learning based methods.

\smallskip
\noindent{\bf Analysis of CNNs:}
Unlike hand-crafted features such as SIFT~\cite{Lowe_2004} or
HOG~\cite{1467360}, the representation learned by CNNs is not obviously
interpretable. For many tasks the CNN is used as a black-box algorithm where it
is not always clear why the CNN is outperforming previous approaches.
Several recent works have attempted to peer into this box,
to better understand the representations learned by CNNs. P. Fischer {\emph{et al.}}~\cite{fischer2014descriptor} 
compares the learned representation with SIFT in a descriptor matching task.
M.D. Zeiler {\emph{et al.}}~\cite{zeiler2014visualizing} propose several heuristic visualization
techniques for units in the network. J. Long {\emph{et al.}}~\cite{long2014convnets} study the 
effectiveness of convnet activation features for tasks requiring correspondence. 
Recently, B Zhou {\emph{et al.}}~\cite{zhou2014object} present a technique to visualize learned
representations of each unit in the network. Here they focus on a large dataset
of scene images and show that object detection is embedded in the network as a result of
learning. We use variants of this approach to evaluate what our temporal
networks have learned.

\smallskip
\noindent{\bf Visual Data Mining:}
Several visual data mining approaches have been used for tasks such as
unsupervised discovery of object categories in image
collections~\cite{grauman2006unsupervised,payet2010set,faktor2014clustering,sivic2005discovering},
or for finding discriminative parts of actions~\cite{raptis2012discovering}, 
cities~\cite{doersch2012makes}, or objects~\cite{lee_styleAware}. There has
also been related work on discovering localized
attributes~\cite{attributediscovery,duan2012discovering,rastegari2012attribute} 
for improved fine-grained recognition.
Most of these approaches start from existing hand-crafted feature
representations and pre-defined measures of visual similarity, then based on
discovered patterns of features in the data, group visual elements into
discovered entities. The most relevant work is from~\cite{lee_styleAware} 
which go beyond simply detecting recurring visual elements to model 
the stylistic differences between objects over time or
space. Unlike this prior work, we do not use hand-crafted visual
representations. Instead, we use representations learned by CNN models and
achieve significantly improved performance. Additionally, we analyze what the
network has learned, temporally, and in comparison to the existing data mining
approach~\cite{lee_styleAware}.

\begin{figure*}
	\centering
		\includegraphics[width=0.90\textwidth]{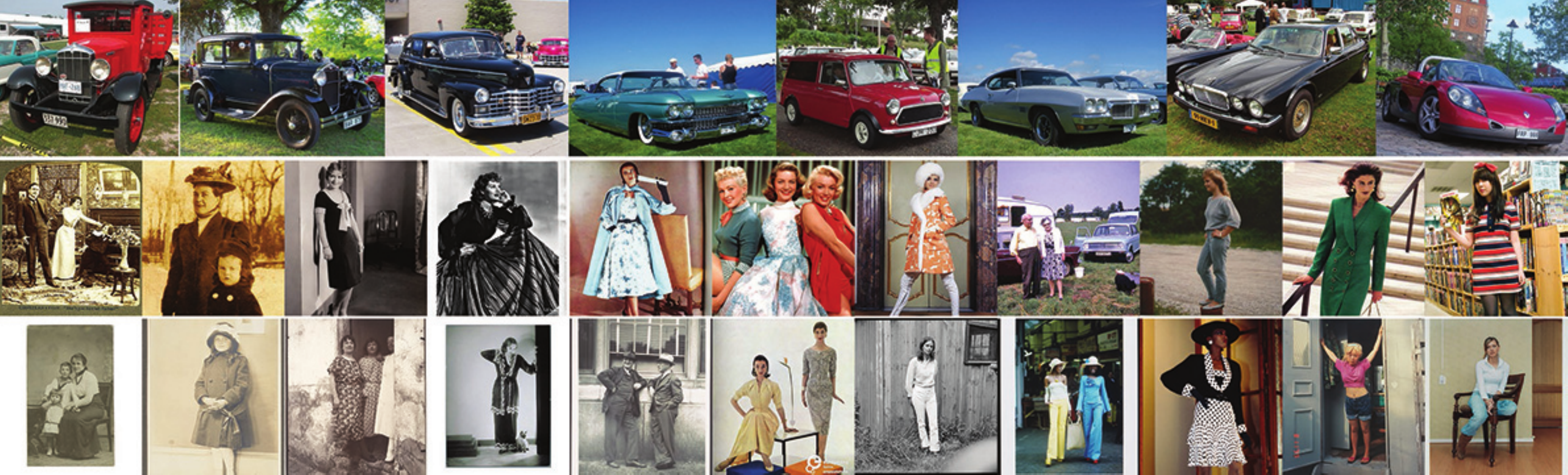}
	\caption{Examples images from CarDb (first row), Flickr clothing dataset (second row) and Museum dataset (third row), sorted by time.}
	\label{fig:dataset}
\end{figure*}


\vspace{-.2cm}
\section{Datasets}
\label{sec:datasets}
We use two main data sources in our work: a) the Car Database (CarDb) collected
by~\cite{lee_styleAware}, and b) a large new collection of clothing
photographs with associated dates, collected from Flickr. 
Since the date information in the Flickr clothing dataset is provided by the user, 
the dates associated with objects can sometimes be noisy.
Therefore, we also collect a smaller dataset of clothing related
photographs from 2 different museum collections. Because these photographs have
been curated by experts, their date labels are quite reliable. We use this
dataset as an alternative test set to evaluate clothing date models trained on
the larger Flickr clothing dataset. As this Museum dataset has a somewhat different
domain than the Flickr images, we can also use it evaluate model generalization, i.e.
training on one dataset and testing on another. 

\smallskip
\noindent {\bf Car Database:}
The Car Database (CarDb)~\cite{lee_styleAware} contains 13,474 photos 
of cars made from 1920 to 1999 resulting 8 temporal classes, 
collected from \url{www.cardatabase.net}. The first row of Figure~\ref{fig:dataset} 
shows example images from CarDb, sorted by time. 

\smallskip
\noindent {\bf Flickr Clothing Dataset:}
We initially collect more than 100,000 clothing related images 
together with their corresponding meta-data from a 
wide variety of 50 Flickr groups focused on vintage fashions, e.g. 
\textit{``Fashions Past - Best and Worst''} and \textit{``As She Was''}. Some of these images 
contain drawings of fashion items or depict other fashion
related images without clear examples of clothing items. To remove these images 
we apply a face detection algorithm~\cite{6755923} 
to automatically filter out images without a depicted person.
The remaining images are manually inspected to remove additional non-photographic content such 
as artwork, painting, and advertisements. 
From the meta-data such as title, description, and tags we automatically
extract a potential decade label such as 90s, 1965, 1954-1957, 1920s etc. 
Then, we quantized these labels into an 11-bin histogram with dates ranging from 1900-2009. 
Finally, a dataset contains 58,350 clothing photographs 
with corresponding meta-data, including photo id, user id, title, description,
tags, longitude, latitude, number of views and groups, etc. Some example
images from our new Flickr clothing dataset, sorted by time, are shown in the 
second row of Figure~\ref{fig:dataset}. 

\smallskip
\noindent {\bf Museum Dataset:}
As the temporal labels of the Flickr dataset have been associated with
images by Flickr users, we also collect an additional dataset that has been
labeled by expert museum curators from 2 different museums; 
the Metropolitan Museum of Art, and Europeana Fashion. 
Europeana Fashion is a museum network co-funded by 22
partners from 12 European countries which represent leading European
institutions and some of the largest collections in the fashion domain. 
The Museum dataset has 9,421 images between 1900 to 2009, 
showing clothing worn on people. We use this dataset as a test set for models trained
on the larger Flickr clothing dataset. Examples of clothing images from the
Museum dataset, sorted by time, are shown in the third row of
Figure~\ref{fig:dataset}.


\vspace{-.3cm}
\section{Approaches}
\label{sec:approaches}

We pursue two approaches for temporal estimation. First, we evaluate
classifiers trained on features from a pre-trained network. Then, we
explore adapting the network to directly predict time period, fine-tuning for this task.

\smallskip
\noindent{\bf Pre-trained models:}
We start with two Convolutional Neural Network (CNN) models pre-trained on 1.2
million labeled images from ImageNet. The first network, AlexNet, was originally described 
by~\cite{NIPS2012_4824} and the latter, VGG, was described by~\cite{simonyan15}, 
both networks are implemented as part of the Caffe framework~\cite{jia2014caffe}. 
For each network, we extract the learned representation from the second fully-connected 
layer and use this 4096 dimensional vector as our visual representation. We experiment with two different classification methods: a linear 
Support Vector Machines (SVM)~\cite{REF08a} with fixed $C_{svm} = 0.1$, and Support Vector Regressors (SVR)~\cite{CC01a} with 
fixed $\epsilon = 0.1$ and set $C_{svr} = 100$. 


\smallskip
\noindent{\bf Fine-tuned models:}
Network fine-tuning has proven successful for adapting networks to new tasks 
beyond their original purpose such as object detection~\cite{girshick2014rcnn}, 
pose estimation and action detection~\cite{poseactionrcnn}, 
or fine-grained category detection~\cite{zhang14finegrained}. In this work, 
we are interested in fine-tuning the original object classification 
model for the temporal estimation task.

From each pre-trained model, described by~\cite{NIPS2012_4824,simonyan15}, we fine-tune 3 different models.
The first model, the fine-tuned Car model, is fine-tuned on 10,130 training images 
and tested on 3,343 images from the CarDb dataset (same train/test split as~\cite{lee_styleAware}).
The second model, the fine-tuned Clothing model, uses $3/4$ of 
the images from the Flickr clothing dataset as training set.
Moreover, since our datasets depict vintage 
photographs from 1900-2009, historic color cues, e.g black and white color 
or sepia tones in early photos and color photographs in later photos, 
appear in both datasets. To evaluate performance without any color cues, 
we also fine-tune a third network using only black/white images from the 
Flickr clothing dataset. Finally, both models are evaluated during 
testing on the portion of the Flickr clothing dataset 
and on the Museum dataset.
To fine-tune each model, we change the output of the last fully connected layer
from 1000 classes to 11 temporal classes (one per decade) for the Flickr clothing dataset and 8 temporal classes for CarDb. We trained our models using stochastic gradient descent with a batch size of 50 examples, momentum of 0.9, weight decay of 0.0005 and decrease the learning rate of the models to 0.00001. Finally, we trained the networks for 50,000 cycles.

\begin{table}
	\small
	\centering	
		\begin{tabular}[b]{l*{6}{c}r}
			\cline{2-4}
			& \textbf{CarDb} & \textbf{Clothing} & \textbf{Museum} \\			
					\hline
			Lee {\emph{et al.}}~\cite{lee_styleAware} & 8.56 & 17.74 & 19.56 \\
			\hline
			Palermo {\emph{et al.}}~\cite{dating_colorim} & - & 17.21 & 21.21\\
			\hline			
			alexNet + SVM		& 7.77 & 12.99 & 17.33\\
			alexNet + SVR		& 8.10 & 16.73 & 22.00 \\
            VGG-16 + SVM		& 6.90 & 12.60 & 16.35 \\
			VGG-16 + SVR		& 7.43 & 15.87 & 18.76 \\
			\hline
			alexNet (BW/FT)	& 6.78 & 17.16 & 17.96 \\
			alexNet (FT) 		& 6.17 & 12.88 & 16.43 \\	
            VGG-16 (BW/FT) & 4.27 & 13.66 & 16.40 \\
            VGG-16 (FT) & \textbf{3.97} & \textbf{11.54} & \textbf{14.23} \\
			\hline
		\end{tabular}	
	\caption{The mean absolute error (years) training and testing on CarDb dataset and training on Flickr clothing dataset and testing on held out clothing and Museum dataset.}
    \label{tab:table2}
\end{table}

\vspace{-.1cm}
\section{Temporal Estimation Experiments}
\label{sec:prediction}
We evaluate performance of our models compared to Lee {\emph{et al.}}~\cite{lee_styleAware} and 
F. Palermo {\emph{et al.}}~\cite{dating_colorim}.

\smallskip
\noindent {\bf Pre-trained Model Performance:}
For the pre-trained models, we extract features from last fully-connected layer, then train SVM and 
SVR models to estimate when an object was made. 
We evaluate performance on both the car and clothing datasets. 
Table~\ref{tab:table2} shows mean absolute error (MAE) in years across features and classifiers, 
including comparisons to previous works~\cite{lee_styleAware} on all datasets.
Though we are relying on pre-trained models in this approach, we already
outperform the previous state of the art on both datasets. Moreover, for clothing, 
the results show that even though the domain shift is quite evident (the results on 
the Museum dataset are worse than results on the held out portion of the Flickr 
clothing dataset), the deep learning feature is beneficial for the temporal estimation task.
Evaluations achieve error reductions of $0.95\pm2.47$ years and $3.19\pm2.06$ years on the Museum and Flickr clothing collections respectively compared to the baseline~\cite{lee_styleAware}.

\smallskip
\noindent {\bf Fine-tuned Model Performance:}
The fine-tuned models consistently outperform the pre-trained models on both object categories. 
For cars, the MAE decreases around $2.48\pm0.86$ years. Similar trends apply for clothing, the fine-tuned model 
decrease MAE around $2.86\pm2.42$ years on the Museum dataset and $1.67\pm1.94$ years on the nosier Flickr clothing dataset.
Moreover, the results confirm that the fine-tuned network learns visual elements beyond color by outperforming two color-based baselines. The first baseline, by F. Palermo {\emph{et al.}}~\cite{dating_colorim} which proposed temporally discriminative features related to the evolution of color imaging processes over time, achieves worse MAE on both datasets compare to our fine-tuned networks. Finally, we also compare our approach with a second baseline, where we fine-tune the network using only black/white images from Flickr clothing dataset. The results show that the B/W model achieves about $2.53\pm1.2$ years higher MAE in this task. These results emphasize that even though color is an important clue, our fine-tuned network is able to learn temporally sensitive features of an object beyond color.

\begin{figure}
	\centering
	\begin{subfigure}[b]{0.45\textwidth}
		\includegraphics[width=\textwidth]{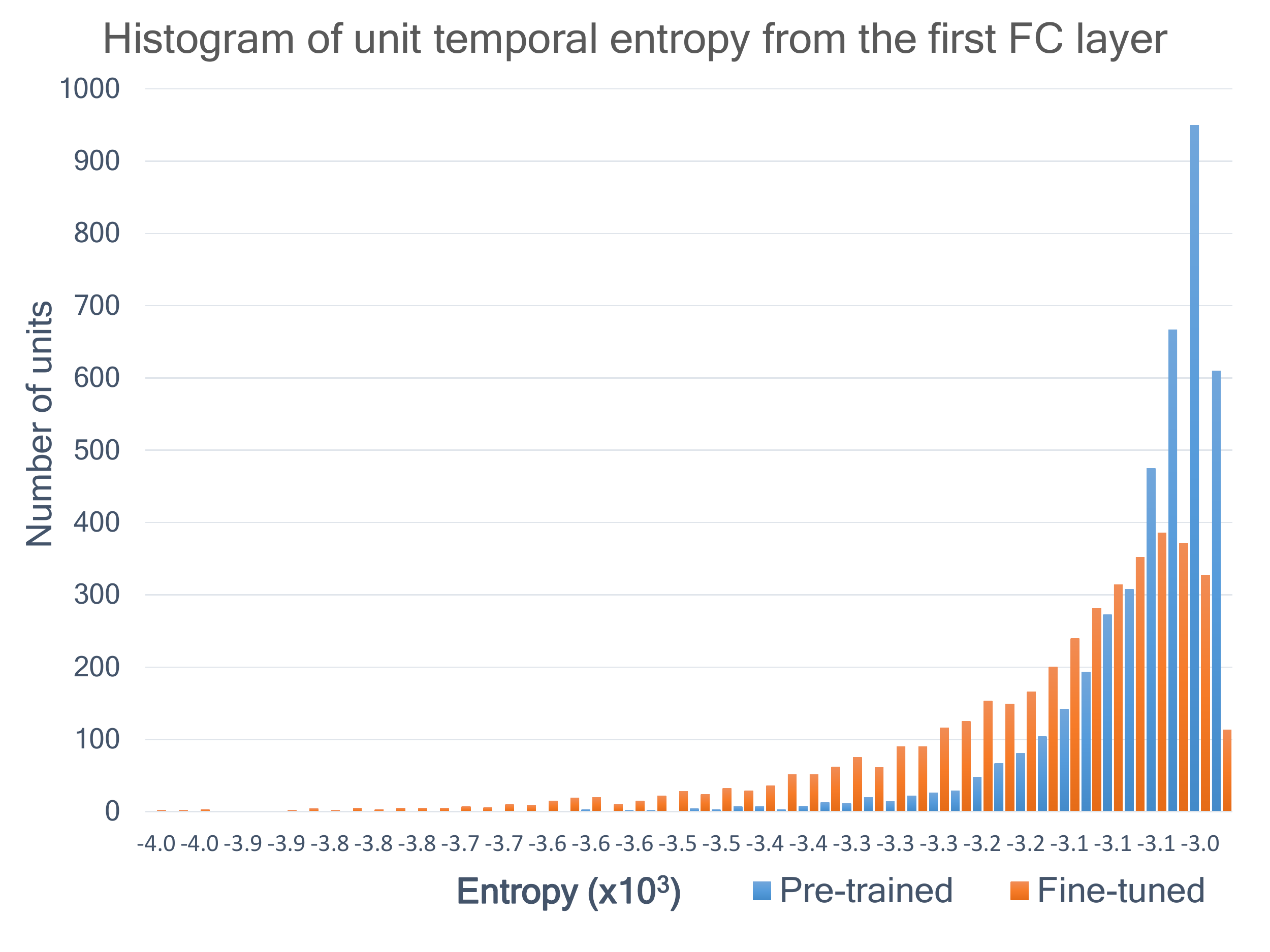}
	\caption{fully connected layer 6.}
	\label{fig:unit_entropy_fc6}
	\end{subfigure} \\
	\centering 
	\begin{subfigure}[b]{0.45\textwidth}
		\includegraphics[width=\textwidth]{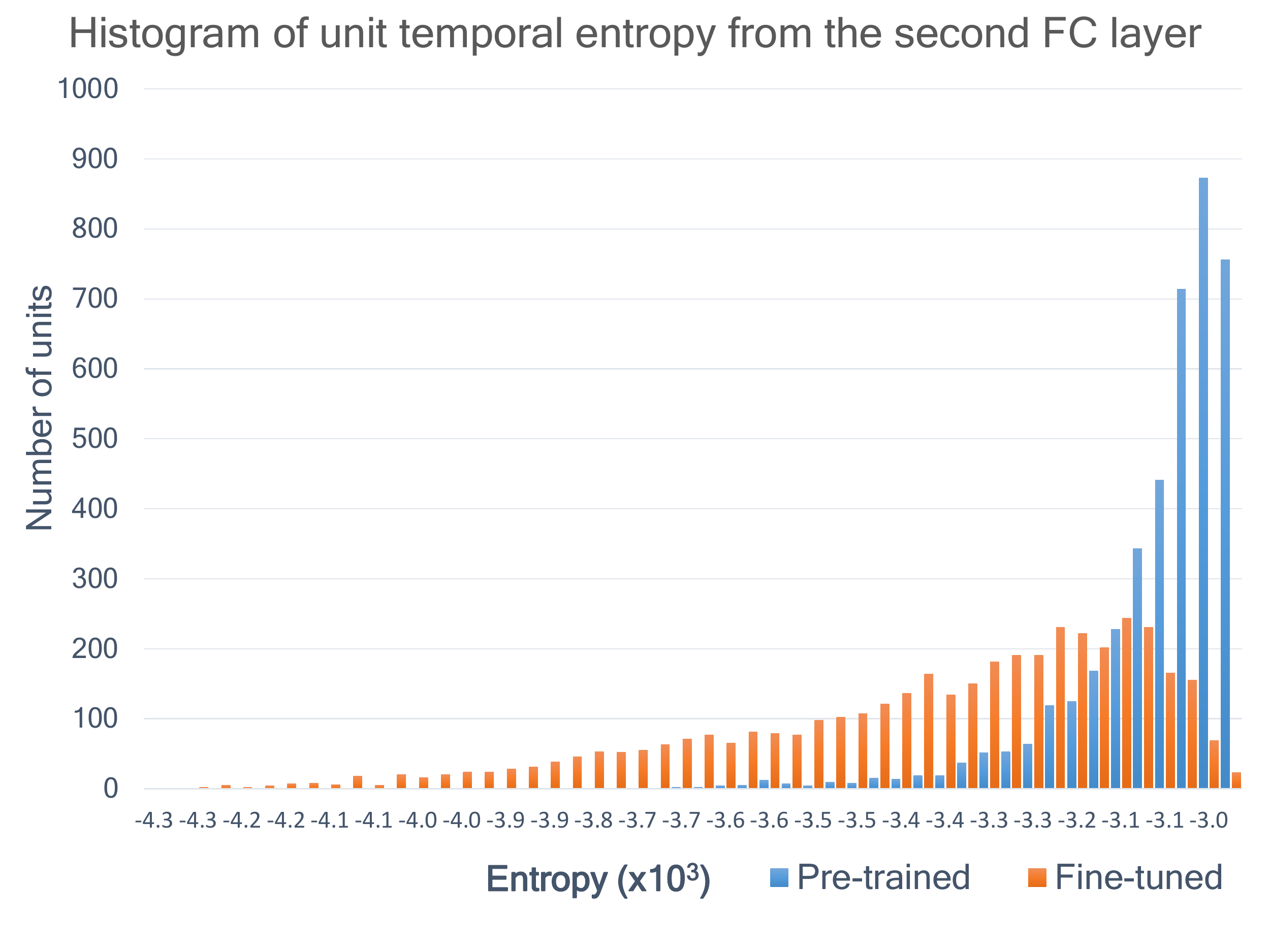}
	\caption{fully connected layer 7.}
	\label{fig:unit_entropy_fc7}
	\end{subfigure}
	\caption{Histogram of unit temporal entropy from 2 different fully connected layers 
	from alexnet(blue) and fine-tuned network(orange).}
\end{figure}

\begin{figure*}
	\centering
			\centering
		\begin{subfigure}[b]{0.22\textwidth}
			\centering
				\includegraphics[width=\textwidth]{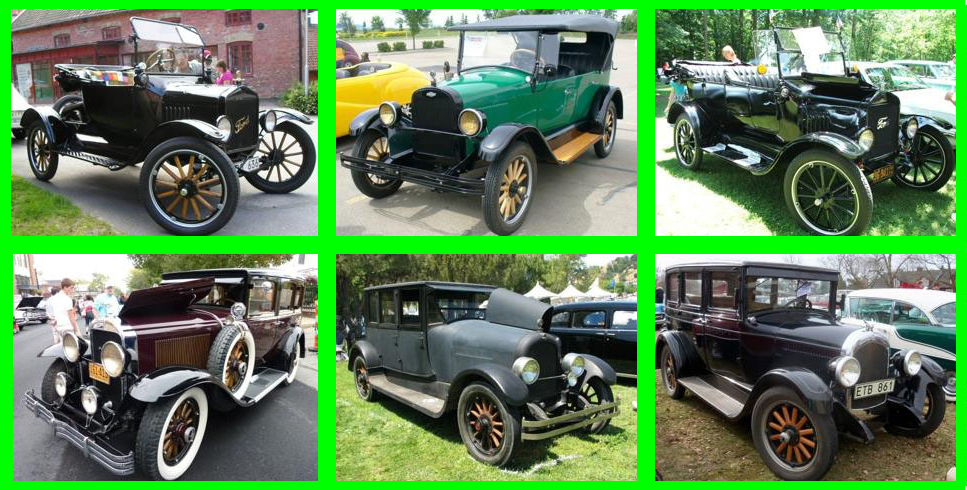}	
				\caption{1920s}
		\end{subfigure}
		\begin{subfigure}[b]{0.22\textwidth}
			\centering
				\includegraphics[width=\textwidth]{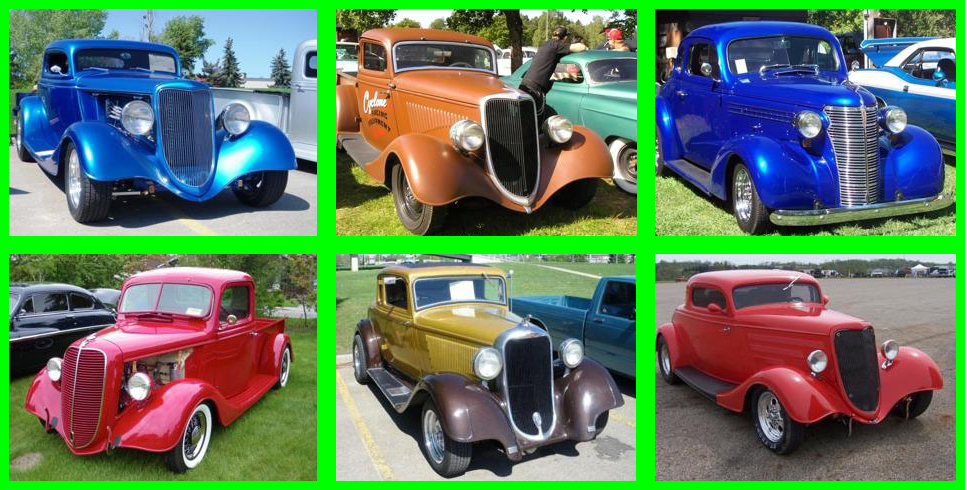}	
				\caption{1930s}
		\end{subfigure}
		\begin{subfigure}[b]{0.22\textwidth}
			\centering
				\includegraphics[width=\textwidth]{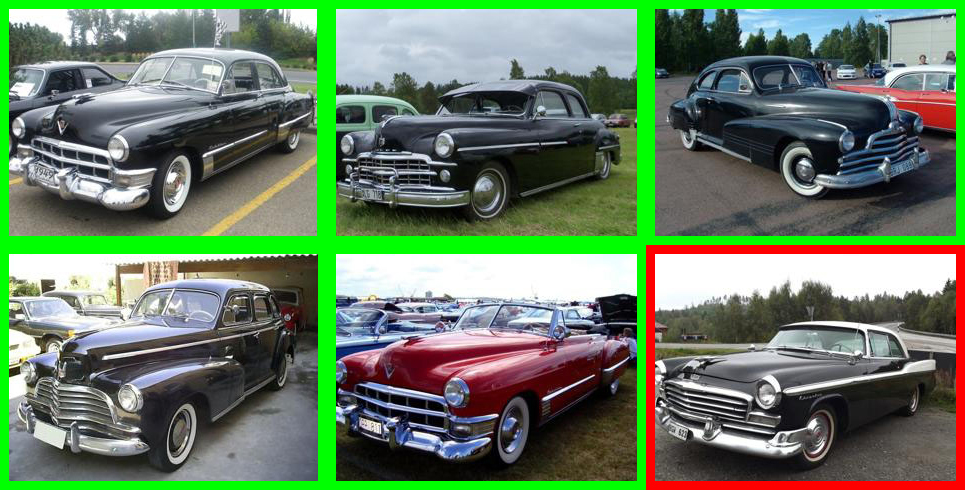}	
				\caption{1940s}
		\end{subfigure}
		\begin{subfigure}[b]{0.22\textwidth}
			\centering
				\includegraphics[width=\textwidth]{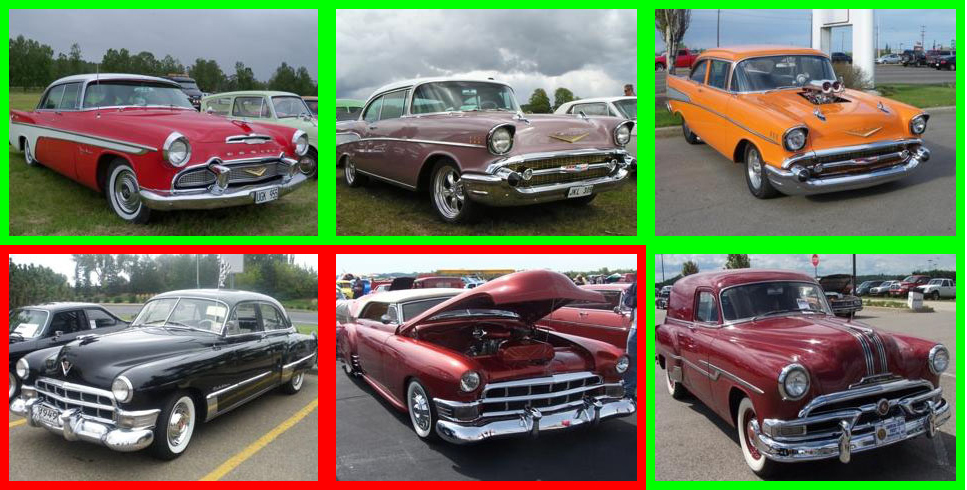}
				\caption{1950s}				
		\end{subfigure}
		\begin{subfigure}[b]{0.22\textwidth}
			\centering
				\includegraphics[width=\textwidth]{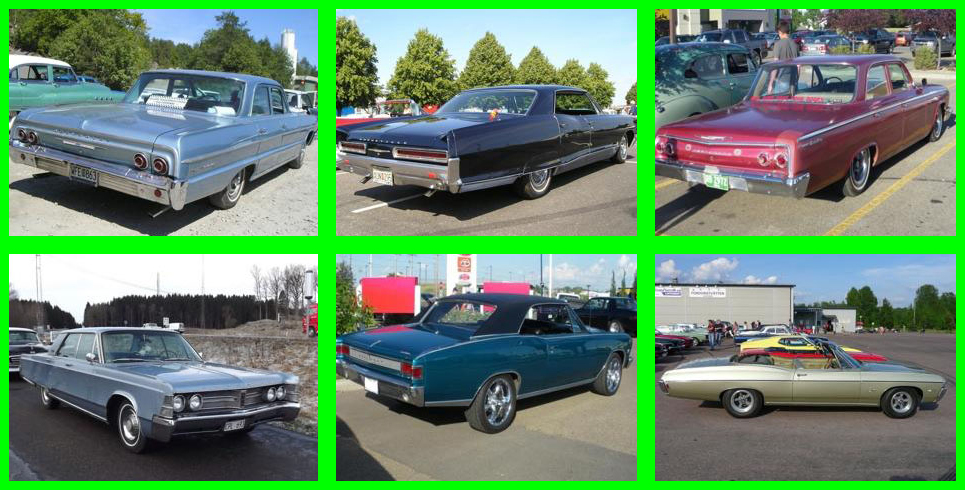}
				\caption{1960s}				
		\end{subfigure}
		\begin{subfigure}[b]{0.22\textwidth}
			\centering
				\includegraphics[width=\textwidth]{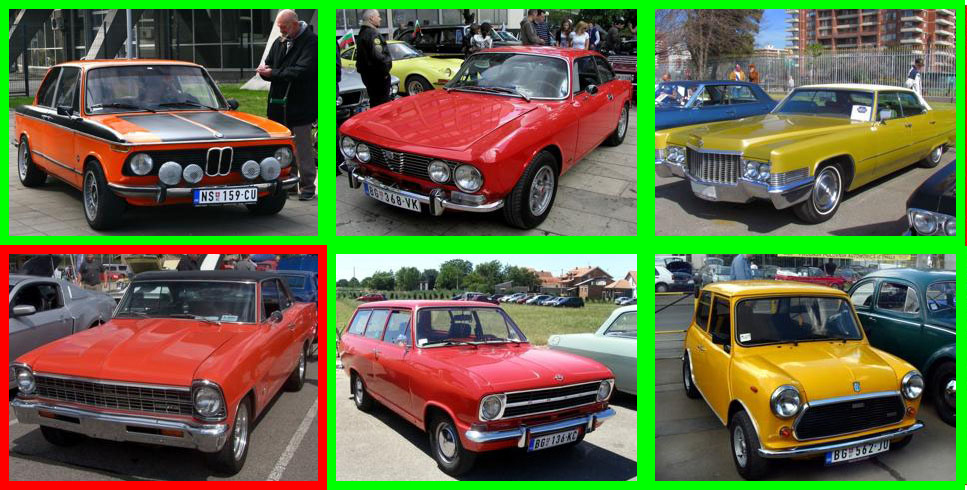}
				\caption{1970s}				
		\end{subfigure} 
		\begin{subfigure}[b]{0.22\textwidth}
			\centering
				\includegraphics[width=\textwidth]{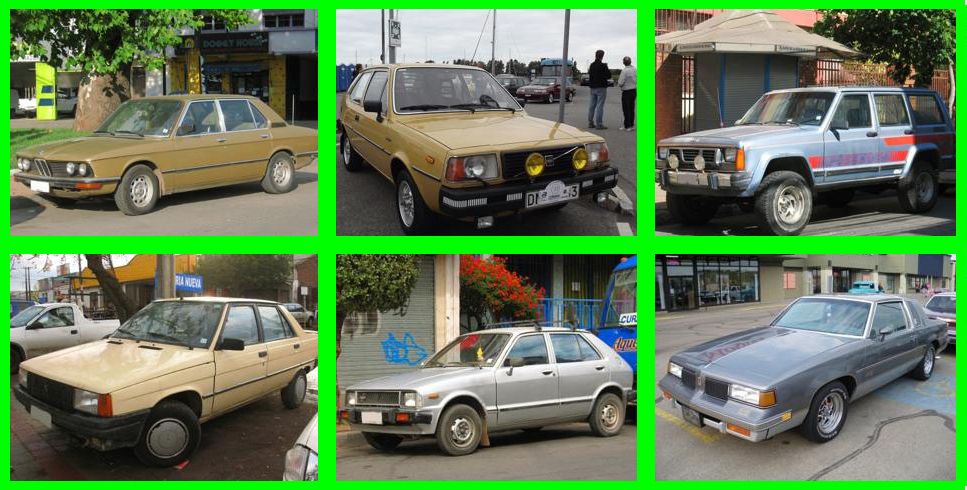}
				\caption{1980s}				
		\end{subfigure}
		\begin{subfigure}[b]{0.22\textwidth}
			\centering
				\includegraphics[width=\textwidth]{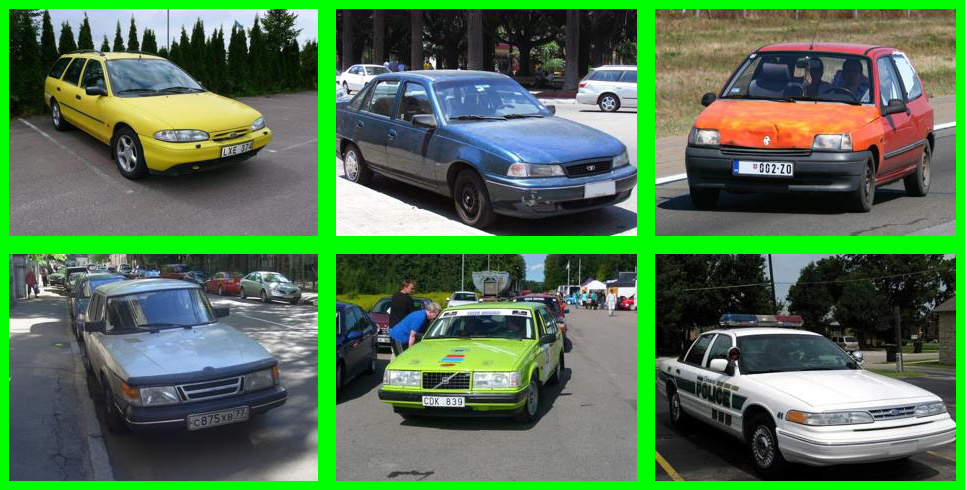}
				\caption{1990s}				
		\end{subfigure}
	\caption{Top 6 images with maximum activation from the low entropy units in each decade. 
	The green rectangle indicates an image is from the same decade as the decade with 
	\textit{the lowest entropy} for a given unit, while the red indicates otherwise.}
	\label{fig:low_entropy_units}
\end{figure*}

\begin{figure*}
	\centering
			\centering
		\begin{subfigure}[b]{0.3\textwidth}
			\centering
				\includegraphics[width=\textwidth]{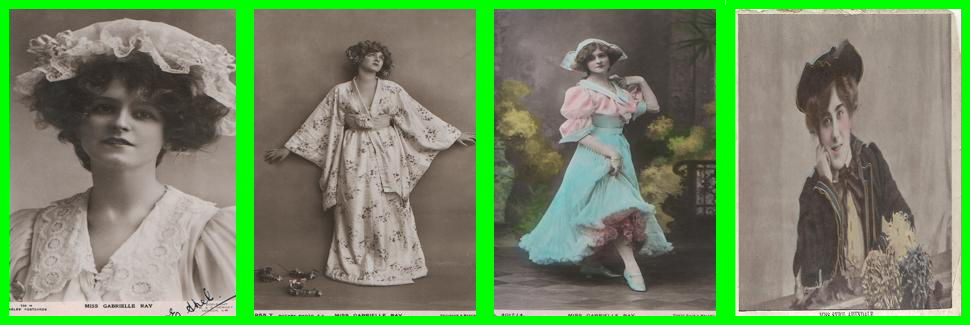}	
				\caption{1900s}
		\end{subfigure}
		\begin{subfigure}[b]{0.3\textwidth}
			\centering
				\includegraphics[width=\textwidth]{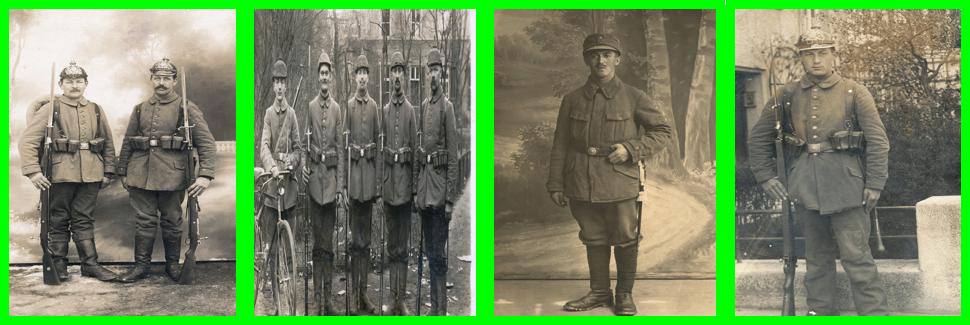}	
				\caption{1910s}
		\end{subfigure}
		\begin{subfigure}[b]{0.3\textwidth}
			\centering
				\includegraphics[width=\textwidth]{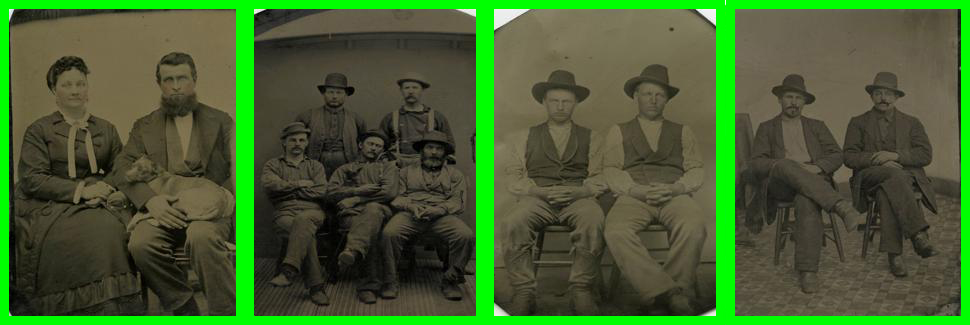}	
				\caption{1930s}
		\end{subfigure} \\
		\begin{subfigure}[b]{0.3\textwidth}
			\centering
				\includegraphics[width=\textwidth]{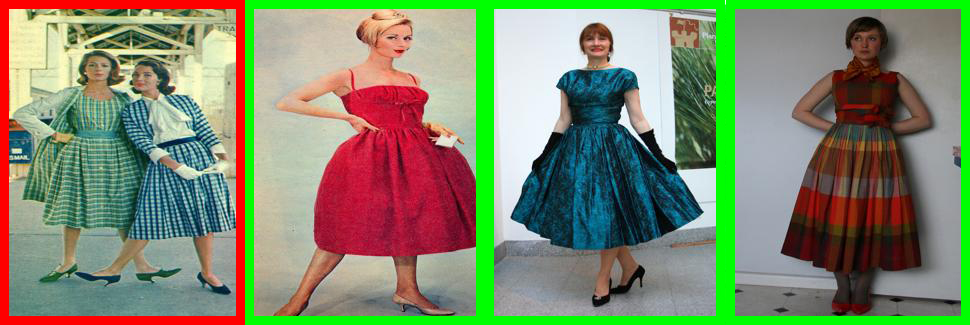}	
				\caption{1950s}
		\end{subfigure}
		\begin{subfigure}[b]{0.3\textwidth}
			\centering
				\includegraphics[width=\textwidth]{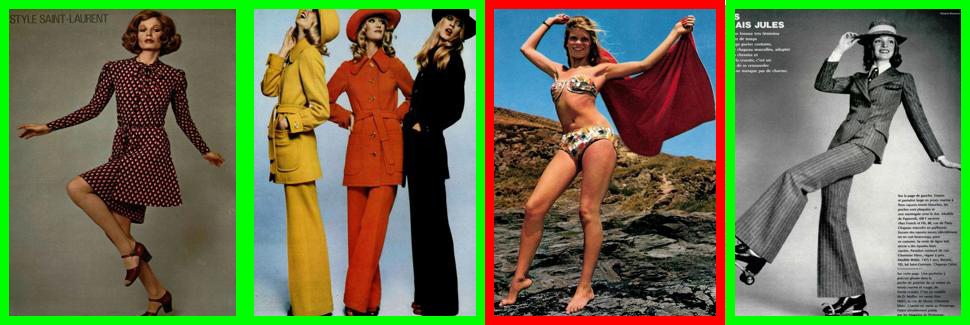}	
				\caption{1970s}
		\end{subfigure}
		\begin{subfigure}[b]{0.3\textwidth}
			\centering
				\includegraphics[width=\textwidth]{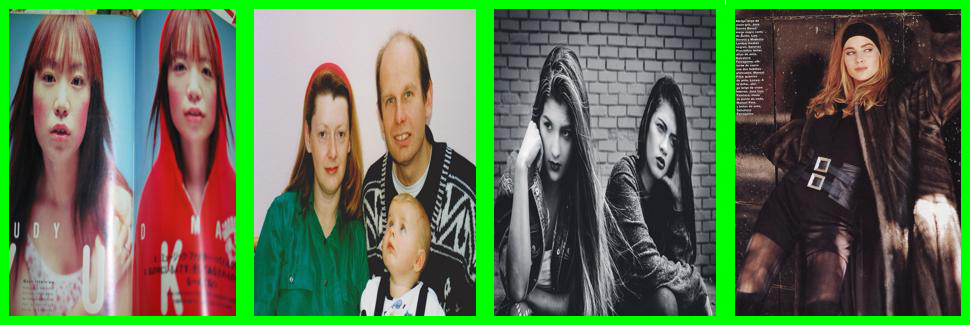}	
				\caption{1990s}
		\end{subfigure} 
	\caption{Top 4 images with maximum activation from the low entropy units in each decade. 
	The green rectangle indicates an image is from the same decade as the decade with 
	\textit{the lowest entropy} for a given unit, while the red indicates otherwise.}
	\label{fig:low_entropy_units2}
\end{figure*}

\vspace{-.1cm}
\section{Deep Network Analyses}
\label{sec:analyses}
Based on these quantitative results, we find that deep learning methods are
promising for temporal estimation task. However, unlike the patch discovery
methods~\cite{lee_styleAware}, the learned
representations from deep networks are not immediately interpretable. Thus, in this
section, we provide some analyses of the CNN networks to gain
additional understanding about what the fine-tuned networks have learned.

\vspace{-.1cm}
\subsection{Temporally-sensitive units} 
\vspace{-.1cm}
Since the fine-tuned network outperforms the pre-trained network, we
hypothesize that there may be interesting differences in the network before and
after fine-tuning. One potential difference that we investigate is the temporal
sensitivity of units. 
To explore the temporal sensitivity of units in both networks, for each unit,
images are ranked by their maximum activation. Then we bin the top $N=500$ maximum 
activation images into a temporal histogram, by decade, and compute the 
entropy of each histogram: $E(u) = -\sum_{i=1}^{n}{H(i)} \cdot log_{2} H(i)$ where $H(i)$ denotes the histogram count for bin $i$ and $n$ denotes the number
of quantized label bins. Lower entropy values indicate higher temporal
sensitivity. Finally, we compute the entropy histogram of all units from 2 Fully-Connected layers as shown in Figure~\ref{fig:unit_entropy_fc6} (first FC layer)
and Figure~\ref{fig:unit_entropy_fc7} (second FC layer). Both histograms show 
that the fine-tuned network has more low entropy units than the pre-trained 
network which indicate that units have been fine-tuned to capture a temporally 
discriminative feature for a specific time period. These results are visible 
in both layers, but are more pronounced in second layer, which makes intuitive 
sense since this layer is closest to the end temporal estimation. Qualitative 
examples of top images ranked by their maximum activation from the low entropy units  
are shown in Figure~\ref{fig:low_entropy_units} for the CarDb and Figure~\ref{fig:low_entropy_units2}
for clothing dataset.

\begin{figure*}
	\centering
	\begin{subfigure}[b]{0.22\textwidth}
			\centering
				\includegraphics[width=\textwidth]{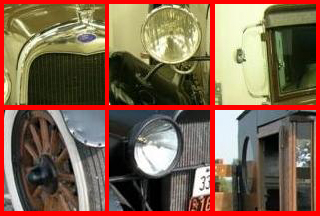}	
				\caption{1920s}
		\end{subfigure}
		\begin{subfigure}[b]{0.22\textwidth}
			\centering
				\includegraphics[width=\textwidth]{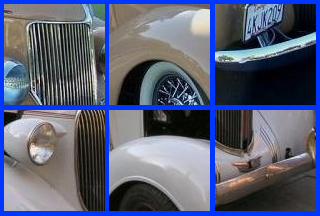}	
				\caption{1930s}
		\end{subfigure}
		\begin{subfigure}[b]{0.22\textwidth}
			\centering
				\includegraphics[width=\textwidth]{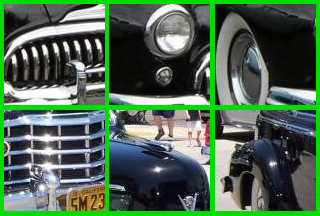}	
				\caption{1940s}
		\end{subfigure}
		\begin{subfigure}[b]{0.22\textwidth}
			\centering
				\includegraphics[width=\textwidth]{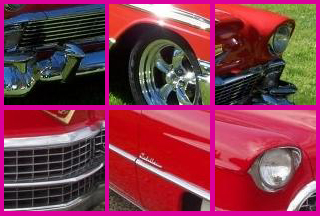}	
				\caption{1950s}
		\end{subfigure}
		\begin{subfigure}[b]{0.22\textwidth}
			\centering
				\includegraphics[width=\textwidth]{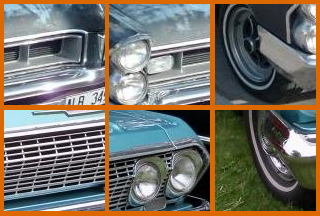}	
				\caption{1960s}
		\end{subfigure}
		\begin{subfigure}[b]{0.22\textwidth}
			\centering
				\includegraphics[width=\textwidth]{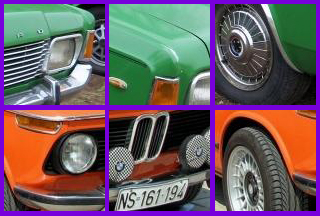}	
				\caption{1970s}
		\end{subfigure}
		\begin{subfigure}[b]{0.22\textwidth}
			\centering
				\includegraphics[width=\textwidth]{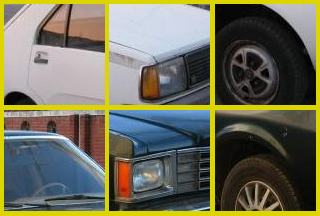}	
				\caption{1980s}
		\end{subfigure}
		\begin{subfigure}[b]{0.22\textwidth}
			\centering
				\includegraphics[width=\textwidth]{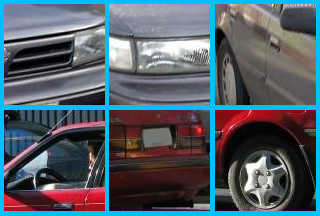}	
				\caption{1990s}
		\end{subfigure}
	\caption{In each rectangle, 3 image regions, from 1920s to 1990s, with the maximum activations from 3 
	different units from fc7 of the fine-tuned network on CarDb are shown.}
	\label{fig:high_activation_car}
\end{figure*}

\begin{figure*}
	\centering
	\begin{subfigure}[b]{0.22\textwidth}
			\centering
				\includegraphics[width=\textwidth]{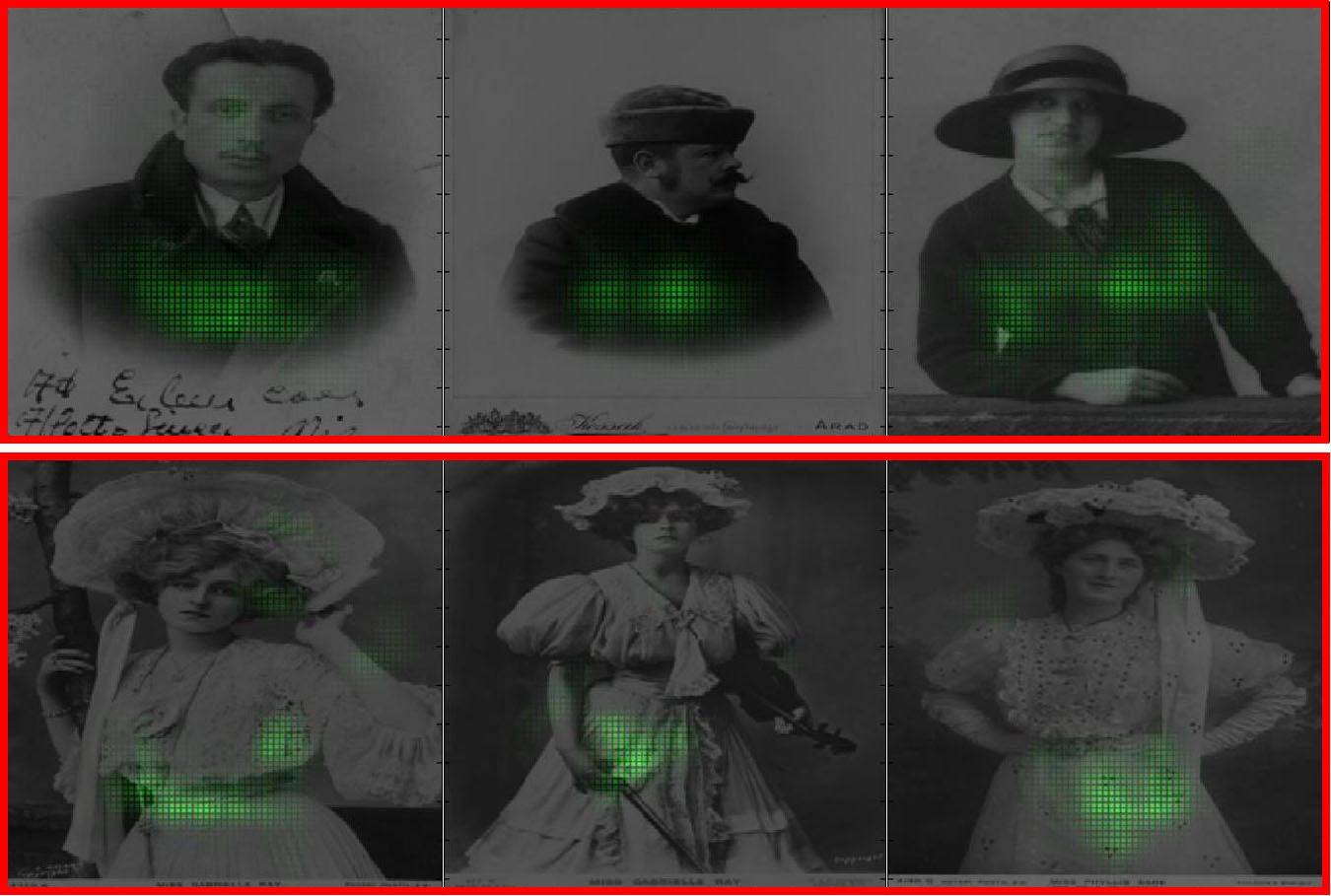}	
				\caption{1900s}
		\end{subfigure}
		\begin{subfigure}[b]{0.22\textwidth}
			\centering
				\includegraphics[width=\textwidth]{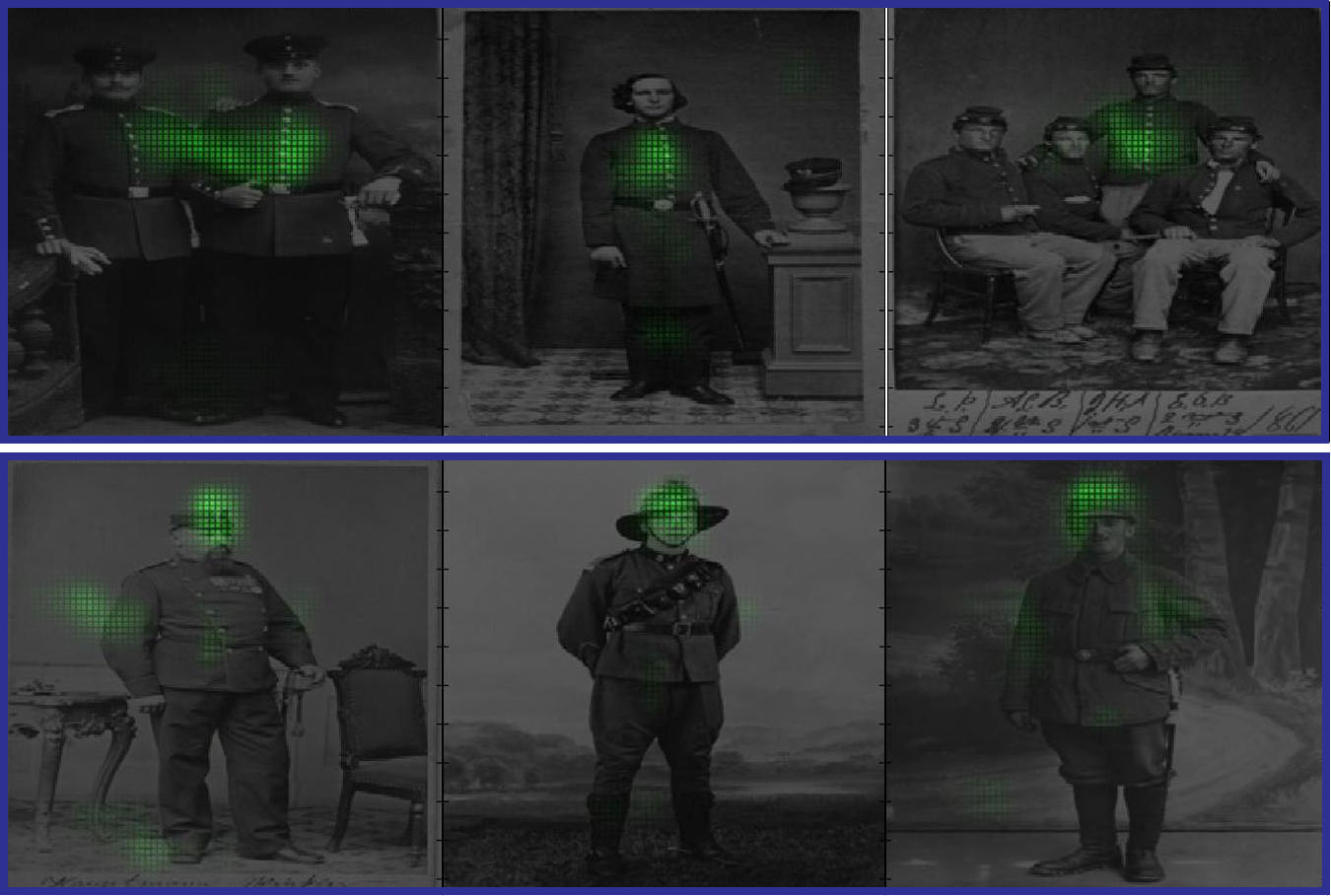}	
				\caption{1910s}
		\end{subfigure}
		\begin{subfigure}[b]{0.22\textwidth}
			\centering
				\includegraphics[width=\textwidth]{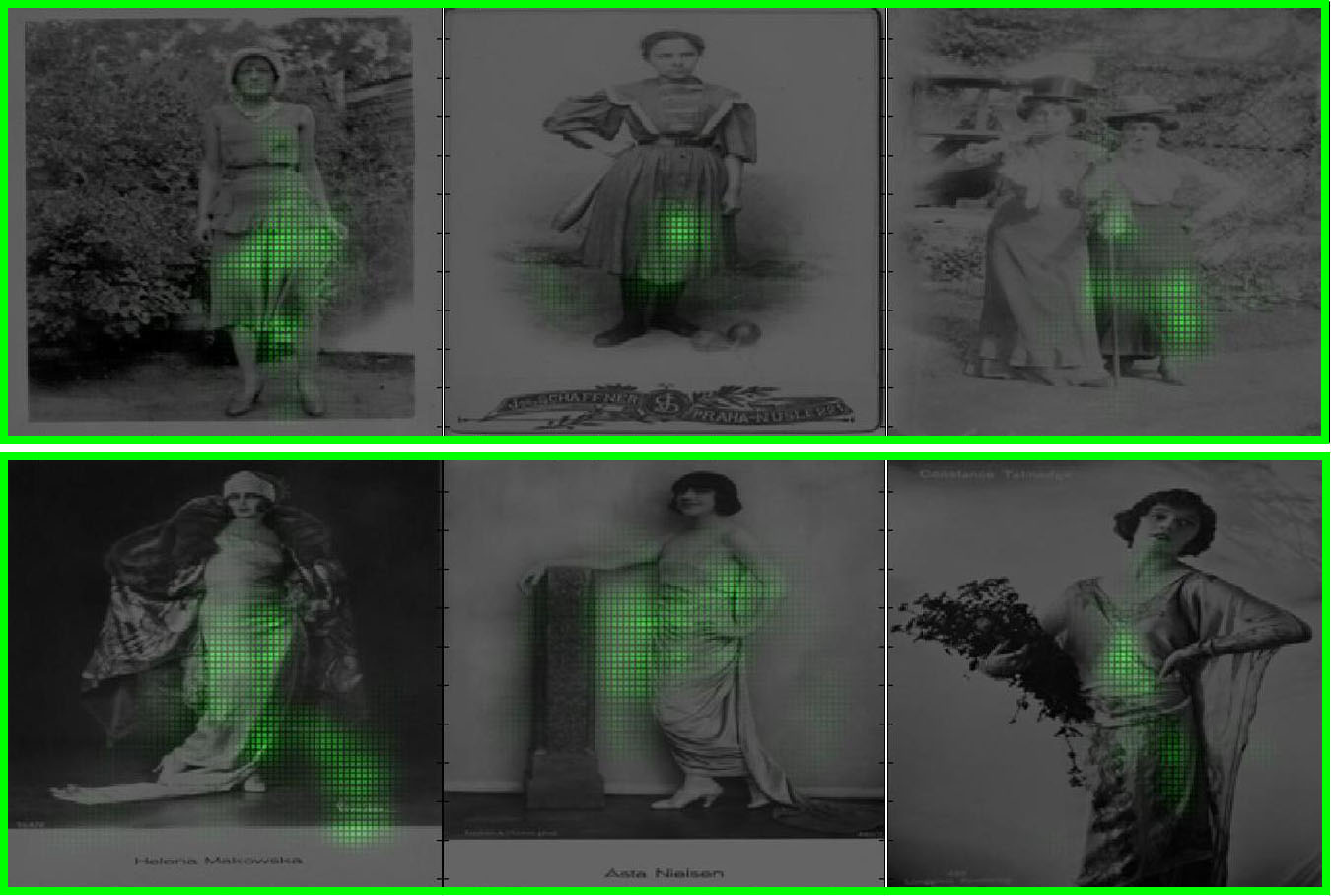}	
				\caption{1920s}
		\end{subfigure}
		\begin{subfigure}[b]{0.22\textwidth}
			\centering
				\includegraphics[width=\textwidth]{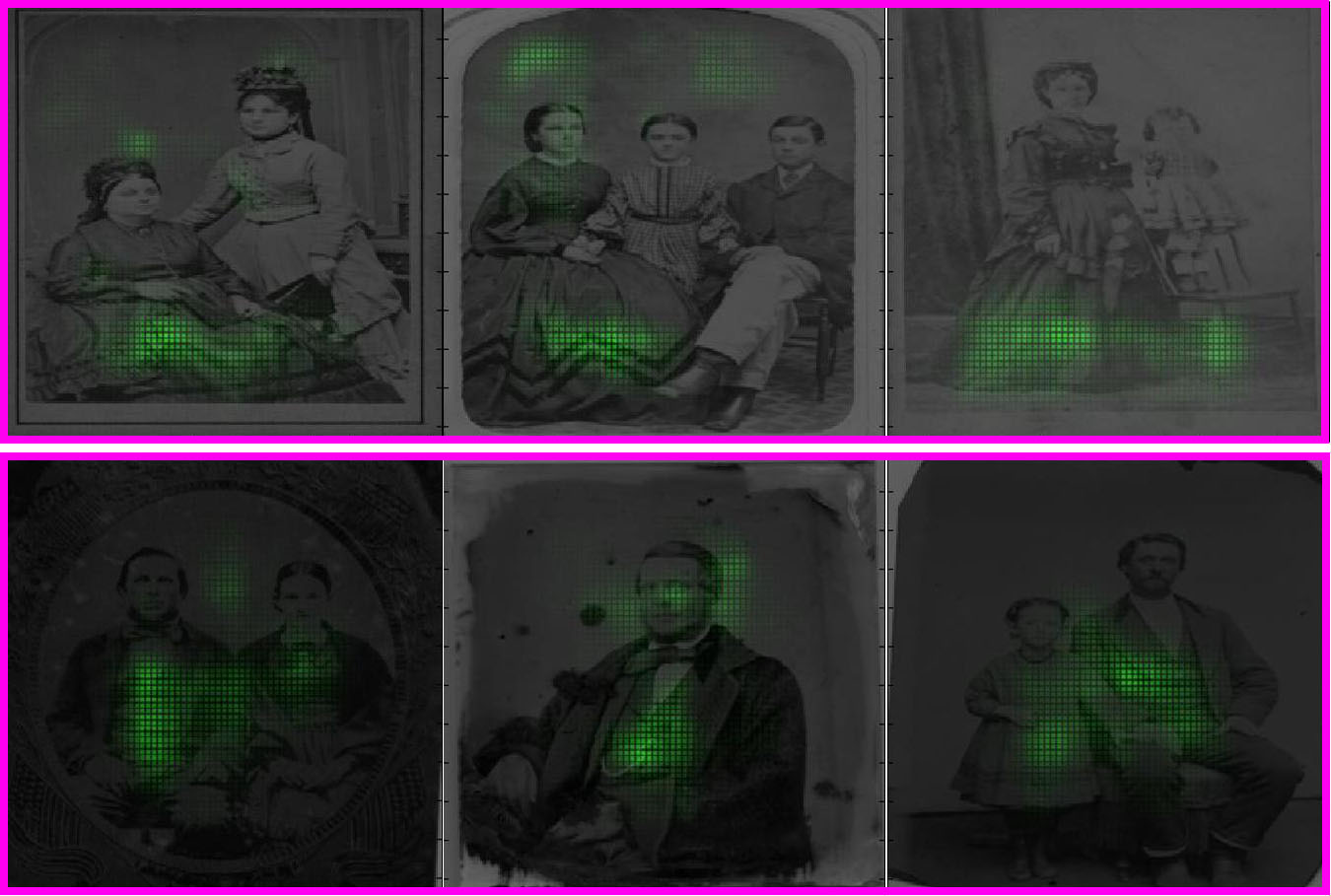}	
				\caption{1930s}
		\end{subfigure}
		\begin{subfigure}[b]{0.22\textwidth}
			\centering
				\includegraphics[width=\textwidth]{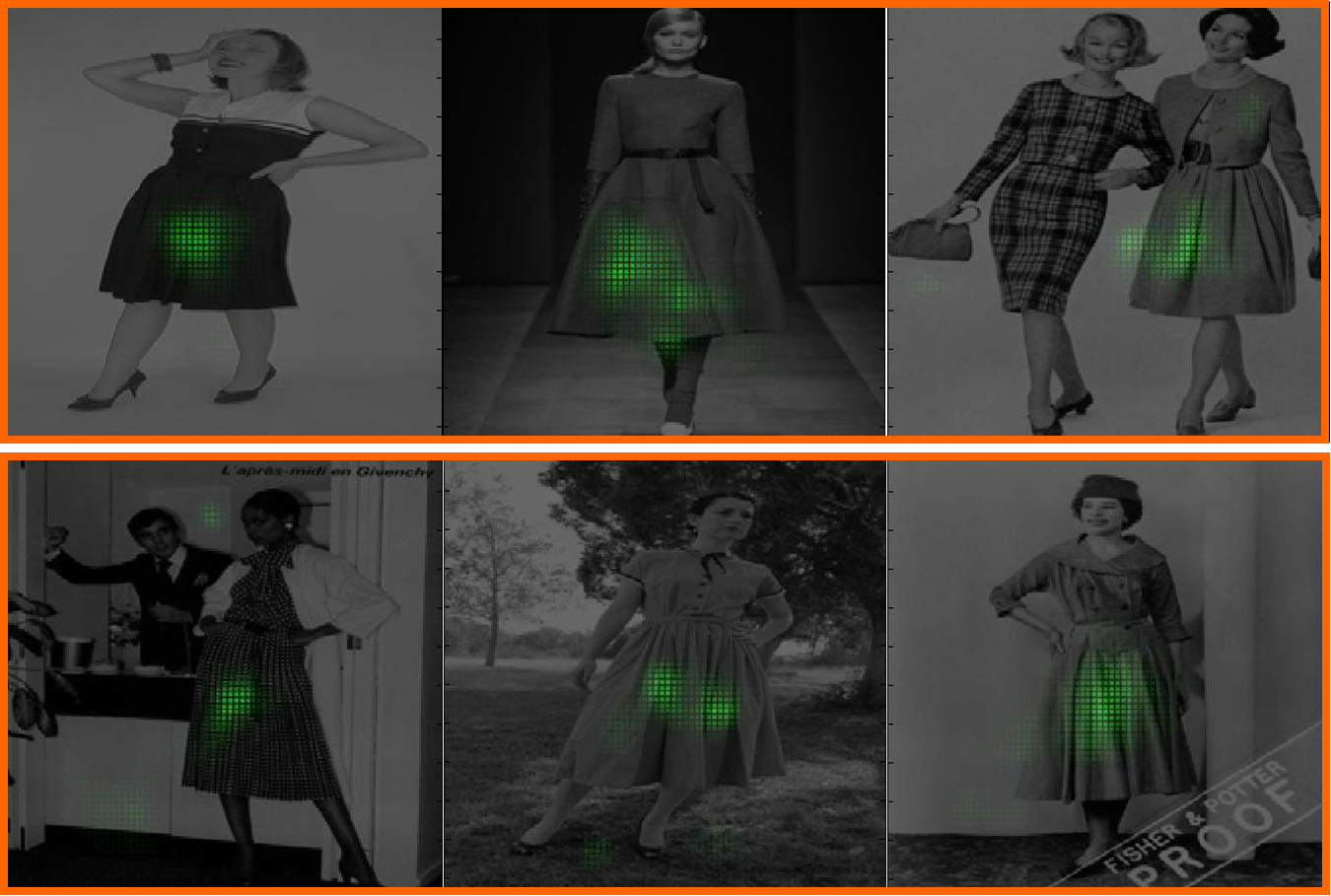}	
				\caption{1940s}
		\end{subfigure}
		\begin{subfigure}[b]{0.22\textwidth}
			\centering
				\includegraphics[width=\textwidth]{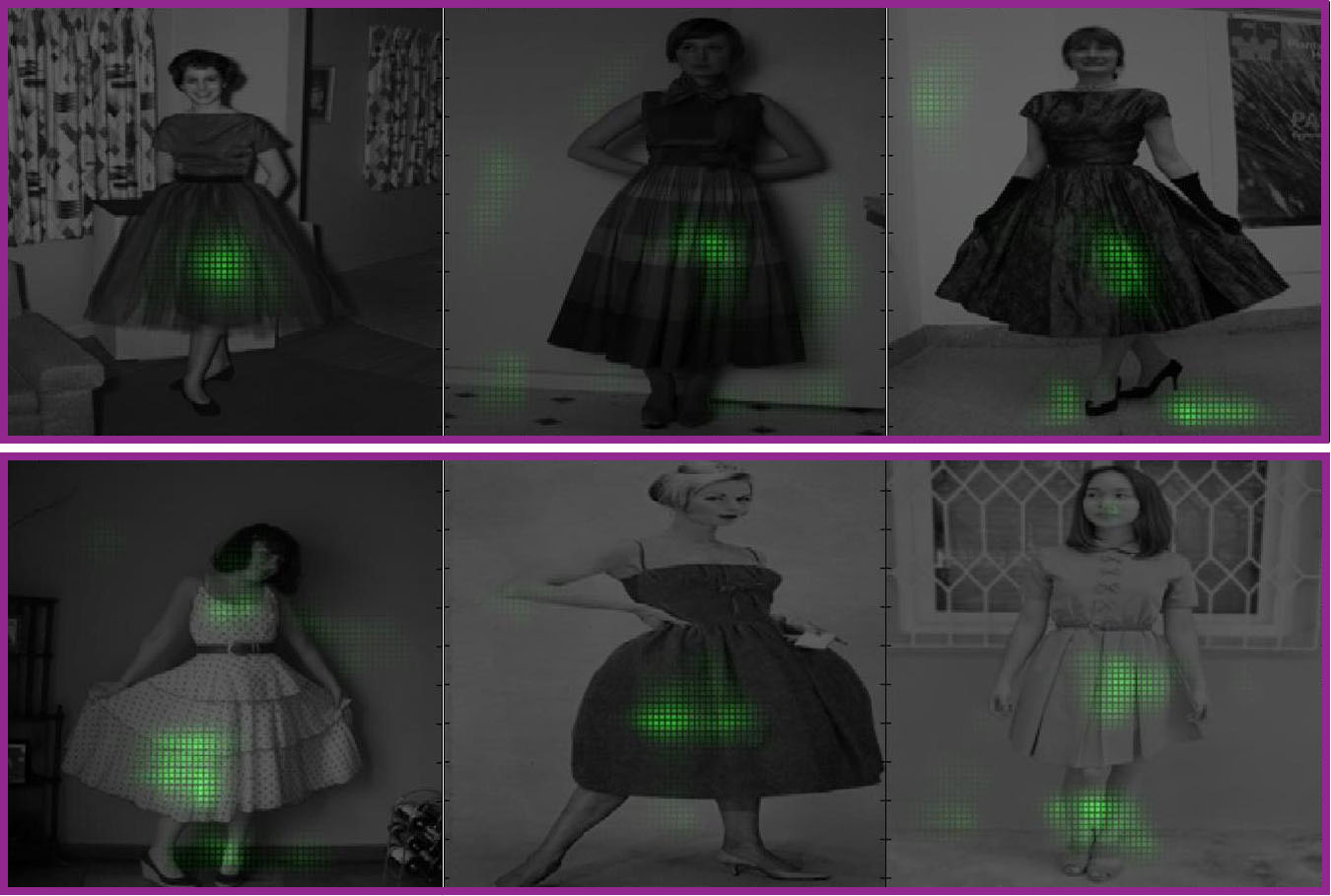}	
				\caption{1950s}
		\end{subfigure}
		\begin{subfigure}[b]{0.22\textwidth}
			\centering
				\includegraphics[width=\textwidth]{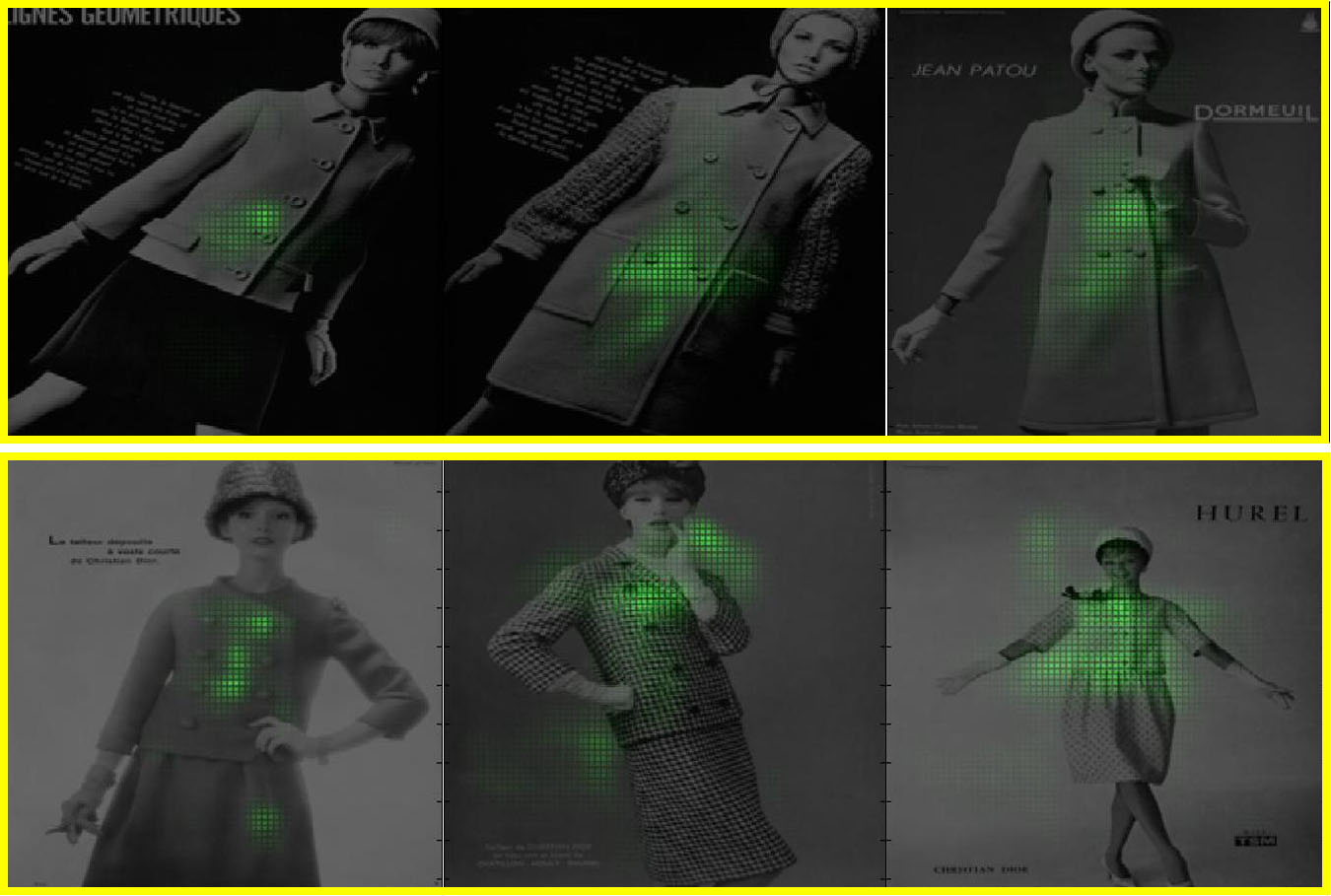}	
				\caption{1960s}
		\end{subfigure}
		\begin{subfigure}[b]{0.22\textwidth}
			\centering
				\includegraphics[width=\textwidth]{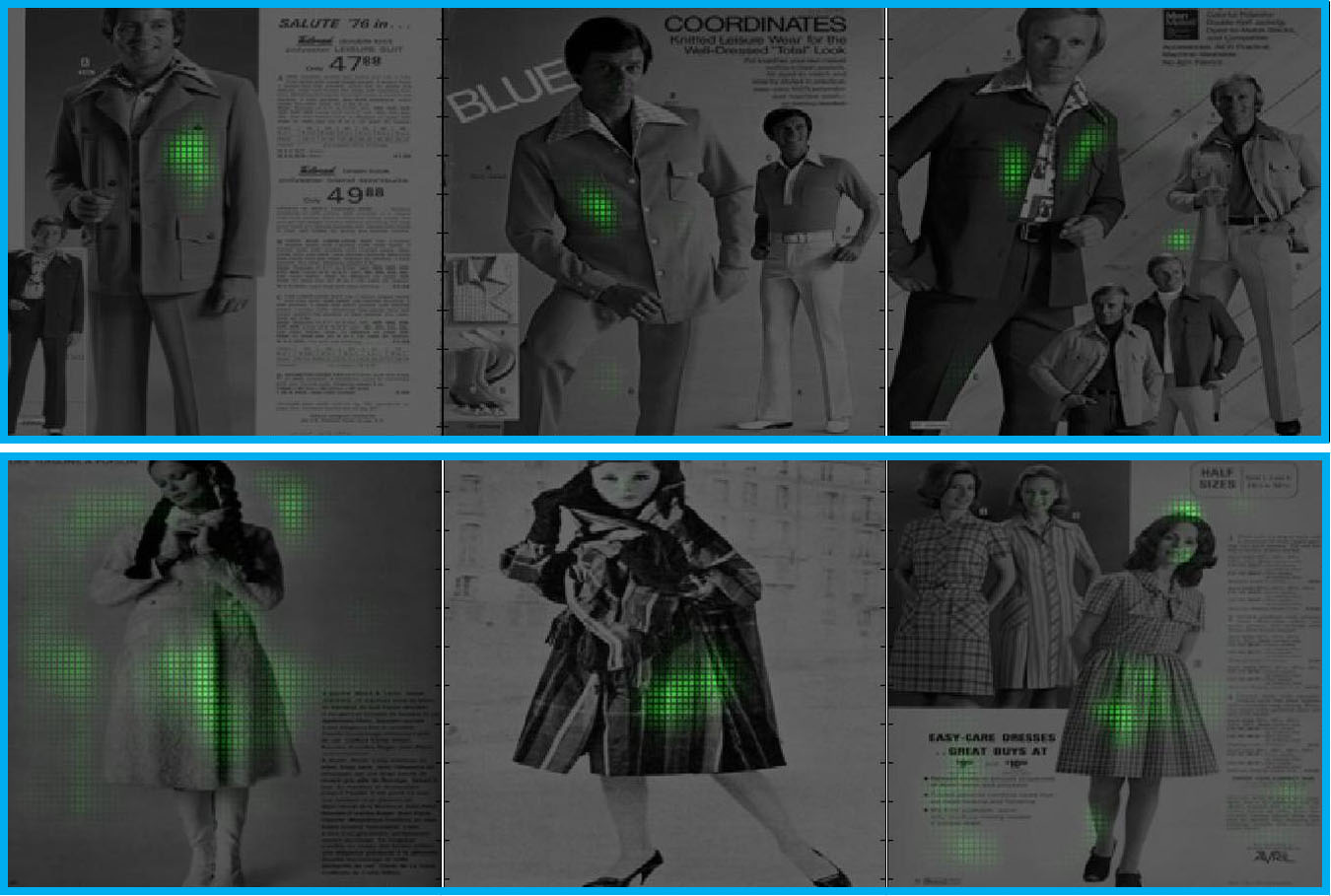}	
				\caption{1970s}
		\end{subfigure}
	\caption{Each rectangle shows 3 images with the maximum activation regions, in green highlight, 
	of units which have the lowest entropy in each decade of the fine-tuned network on Flickr Clothing dataset.}
	\label{fig:high_activation_clothing}
\end{figure*}
\vspace{-.1cm}
\subsection{Unit activation analysis}

Beyond more highly-tuned temporal sensitivity in units, we would like to
understand whether the network has learned to detect temporally
discriminative object appearances or not. Thus, we investigate the 
unit activation patterns to better understand whether the temporally 
sensitive regions correspond to semantic elements of objects. 
To do so, we follow the data-driven approach proposed by ~\cite{zhou2014object} 
to estimate the learned receptive fields (RFs) of units. 
To estimate a unit's RF, images are ranked by their maximum activations 
for that unit, and the top $K$ images are selected to identify image regions that led to these high activations.
To recover the high activation regions within an image, each image is
replicated many times with a small occluder of size 11x11 placed at 
one of about 5,500 locations in a dense grid (stride 3 pixels) in the images. Each occluded image is evaluated by the same network and the change in activation versus the original is calculated. Those differences are combined into a discrepancy map over the image. The intuition behind this approach is that if there is a large discrepancy between activations before and after occlusion, then the occluded region is important for activating that unit.

In order to investigate the most informative regions in an image, 
we focus on image regions which highly contribute to the
prediction decision. Therefore, we first select all true positive
images from the fine-tuned network. For a given image, we rank the
units based on their contribution to the prediction decision and
assign the image to the top $N$ units.  For each unit, we compute the
discrepancy map of assigned images, following~\cite{zhou2014object}. 
Figure~\ref{fig:high_activation_car}-~\ref{fig:high_activation_clothing} 
show image regions that caused the maximum activation for the given
unit from the last FC layer from CarDb and Flickr clothing dataset. 
Results indicate that units in the fine-tuned network respond to temporally 
informative parts such as front bumpers, headlights, or wheels for cars and 
cinched-in waists (40s-50s), mod dresses (60s) and leisure suits (70s) for clothing.

\begin{figure*}
	\centering
		\begin{subfigure}[b]{0.48\textwidth}
			\centering
				\includegraphics[width=0.32\textwidth]{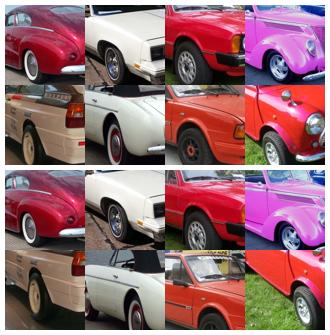}	
				\includegraphics[width=0.32\textwidth]{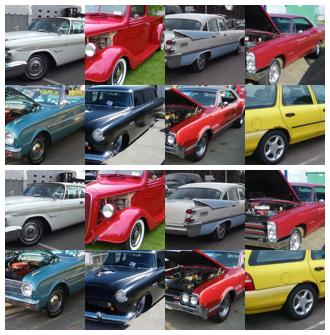}	
				\includegraphics[width=0.32\textwidth]{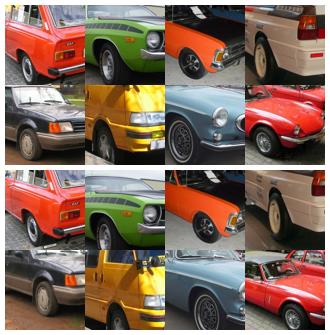}
				\caption{high correlation}
				\label{fig:high_correlation_a}
		\end{subfigure}
		\begin{subfigure}[b]{0.48\textwidth}
			\centering					
				\includegraphics[width=0.32\textwidth]{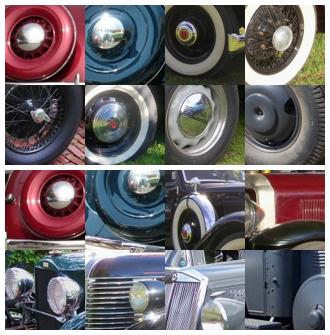}	
				\includegraphics[width=0.32\textwidth]{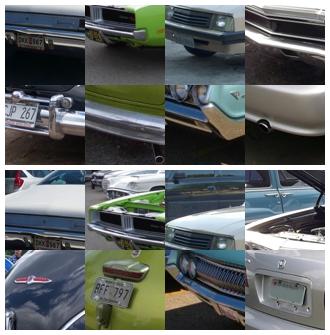}	
				\includegraphics[width=0.32\textwidth]{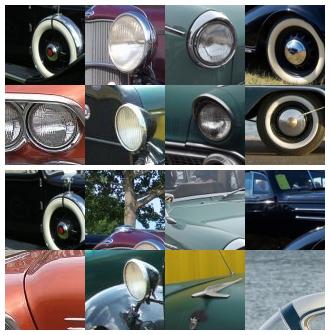}
				\caption{poor correlation}
				\label{fig:high_correlation_b}
		\end{subfigure}\\
	\caption{For each block, the top two rows show style-sensitive patches from~\cite{lee_styleAware}, while the bottom two rows show regions with the maximum activation from the same images. While (a) shows unit activations which are highly correlated to style-sensitive patches, (b) shows unit activations which are poorly correlated to style-sensitive patches}
	\label{fig:high_correlation}
\end{figure*}

\begin{figure}
	\centering
		\includegraphics[width=0.45\textwidth]{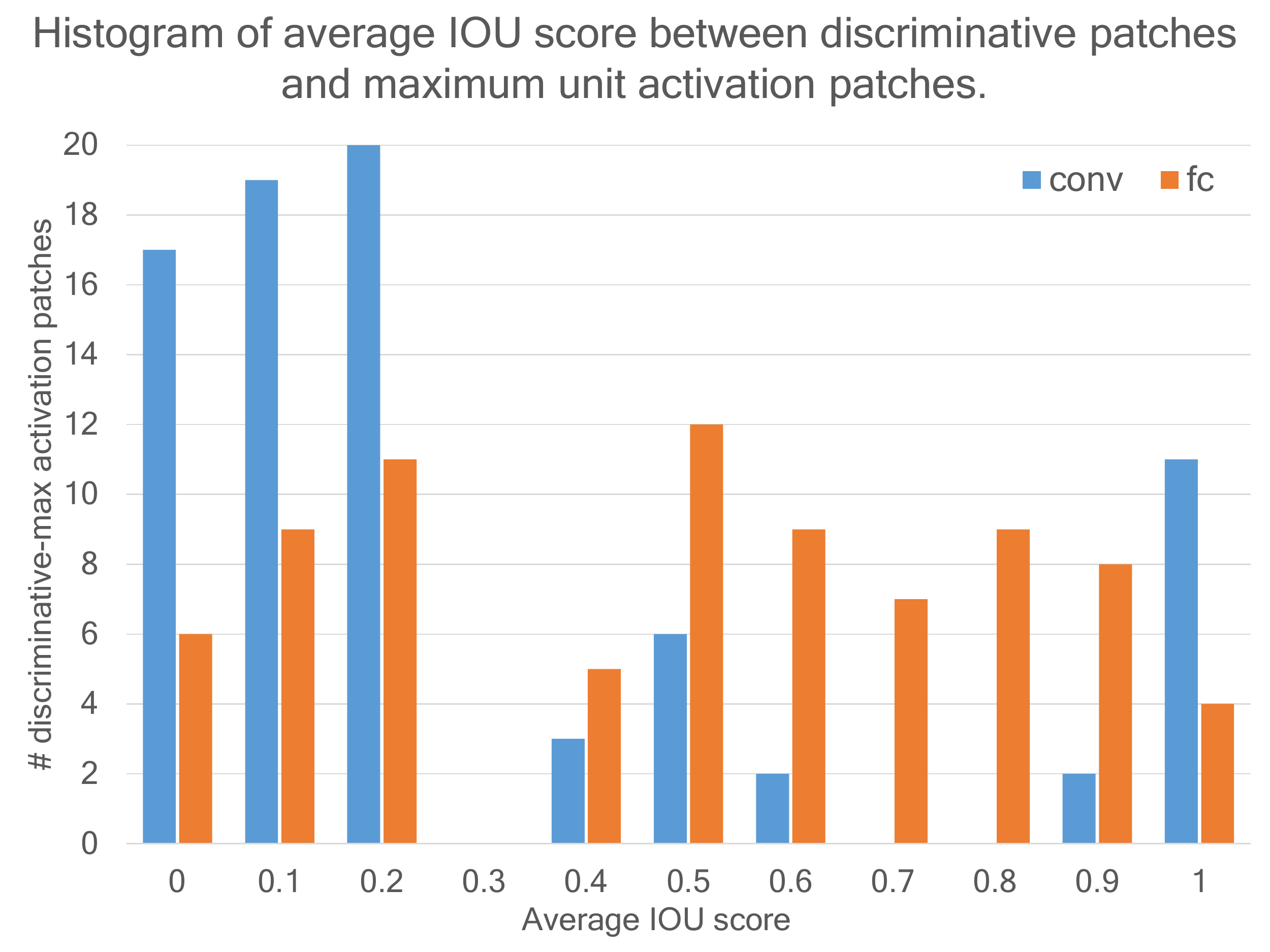}
	\caption{The average IoU score between style-sensitive patches and maximum 
		activation patches.}
	\label{fig:iou_score}
\end{figure}


\vspace{-.1cm}
\subsection{Discriminative part correlation}
So far we have observed temporal sensitivity adaptation and indications that nodes in the network have tuned their activations to particular object parts. These results lead us to new interesting
questions.  Are the parts discovered by the network discriminative in time?  Do
these visual elements correspond to the style-sensitive elements discovered by
~\cite{lee_styleAware}?


To identify correspondences between the visual elements learned by our
fine-tuned network and the style-sensitive elements proposed by~\cite{lee_styleAware}, 
we search for units which have a similar behavior as their style-sensitive detectors.
We look for two behavioral factors: (1) high responses on similar sets of images, 
and (2) similar localization patterns. To implement this, we generate two image rankings; 
the first one is based on the maximum activation for a given unit $u$ over images, 
and the other is based on the maximum detector confidence for a given detector $d$ 
over images. Then, we defined the correlation $C$ between unit $u$ and generic 
detector $d$ as $C(u,d) =\frac{\left|A_n \cap D_n\right|}{n}$ where $A_n$ are 
the set of top $n$ images from the activation based ranking and $D_n$ are the 
set of top $n$ images from the detection based ranking ($n=30\%$ in all experiments). 
For a given detector, we rank all units in each layer by the correlation score.

We evaluate this correlation on units in the last convolutional layer (conv) and a second fully connected layer (fc). The average correlation scores between style-sensitive detectors and their top 5 correlated units from conv and fc are 0.521 and 0.543 respectively, indicating that about
half of the units in both layers overlap with style-sensitive detectors.
Finally we compute the average Intersection-over-Union score (IoU) between 
our maximum activation patches and the style-sensitive patches. More specifically, 
for each correlated unit, we rank images by the maximum activation. Then, 
we sample $80\times80$ pixel patches with maximum activation and compute the 
IoU score between this patch and the style-sensitive patch from the correlated detector. 
In this experiment, we sample a maximum activation patch from the top 20 images/unit with 5 correlated units/style-sensitive detector. The average 
IoU scores of $20\times5$ style-sensitive and maximum activation patches are shown in Figure~\ref{fig:iou_score}. 

These results, again, emphasize that units in the network are 
fine-tuned to temporally sensitive parts of an object. Additionally, we find that
only 7.5\% of the patches have average IoU score $< 0.1$ 
while 61.25\% of the patches have the average IoU score $\geq 0.5$, confirming
that the style-sensitive parts from ~\cite{lee_styleAware} are automatically 
discovered by our network. Qualitative examples of high/poor correlation 
patches are shown in Figure~\ref{fig:high_correlation}. While Figure~\ref{fig:high_correlation_a} 
shows the maximum activation patches from units which are highly correlated 
with style-sensitive parts proposed by~\cite{lee_styleAware}, Figure~\ref{fig:high_correlation_b} 
shows visual elements which have low correlation to~\cite{lee_styleAware}.  
We posit that these additional patches are also temporally sensitive, 
and contribute to our improved performance. 

\begin{figure*}
	\centering     
	\begin{subfigure}[b]{0.32\textwidth}
		\centering
			\includegraphics[width=\textwidth]{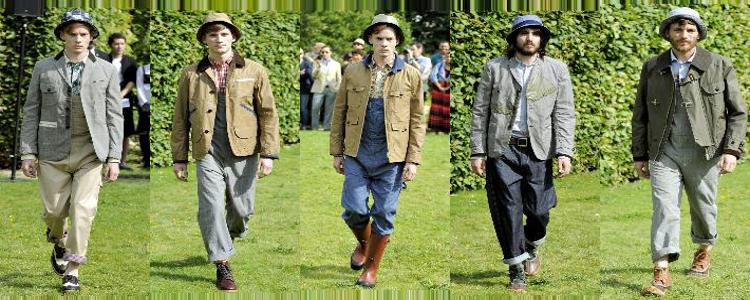}	
			\caption{1940s}
	\end{subfigure}
	\begin{subfigure}[b]{0.32\textwidth}
		\centering
			\includegraphics[width=\textwidth]{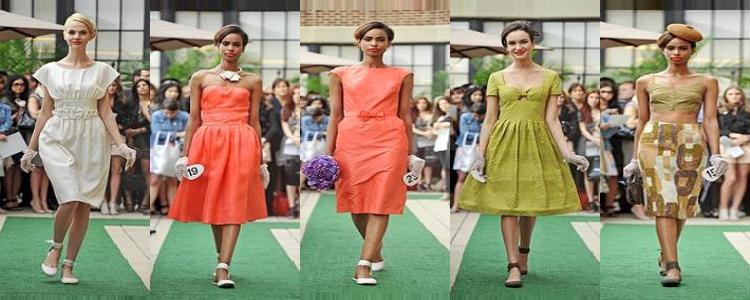}	
			\caption{1960s}
	\end{subfigure}
	\begin{subfigure}[b]{0.32\textwidth}
		\centering
			\includegraphics[width=\textwidth]{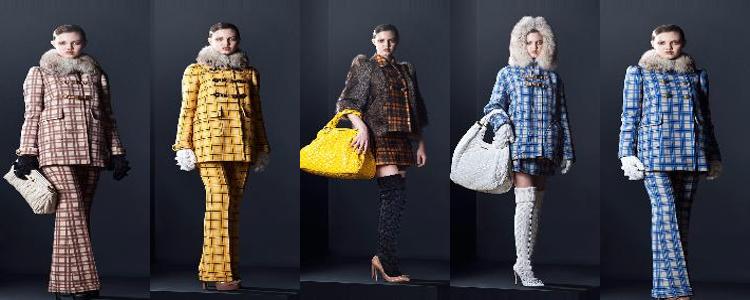}	
			\caption{1970s}
	\end{subfigure}
	\begin{subfigure}[b]{0.32\textwidth}
		\centering
			\includegraphics[width=\textwidth]{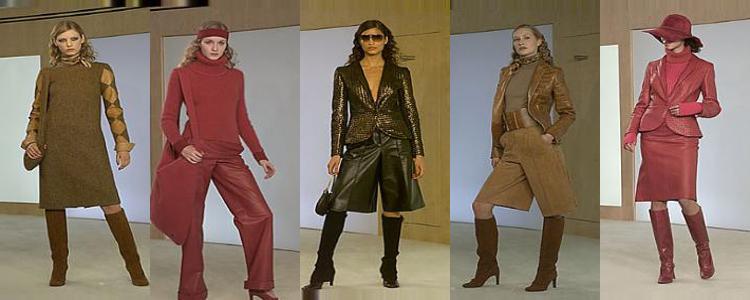}	
			\caption{1970s}
	\end{subfigure}
	\begin{subfigure}[b]{0.32\textwidth}
		\centering
			\includegraphics[width=\textwidth]{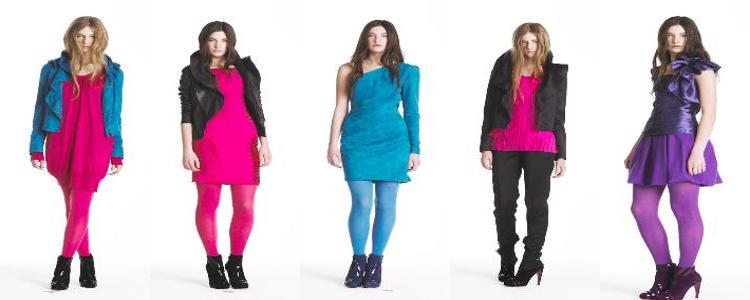}	
			\caption{1980s}
	\end{subfigure}
	\begin{subfigure}[b]{0.32\textwidth}
		\centering
			\includegraphics[width=\textwidth]{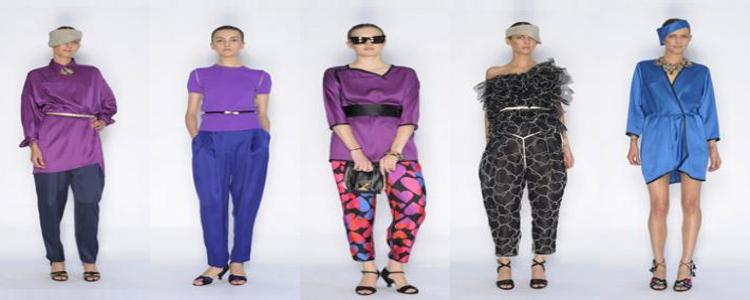}	
			\caption{1980s}
	\end{subfigure}
\caption{Predicting vintage influence in fashion collections. (a)-(f) indicate the decade of predicted influence.}
\label{fig:app}
\end{figure*}

\begin{figure*}
\centering     

\begin{subfigure}[b]{0.32\textwidth}
		\centering
			\includegraphics[width=\textwidth]{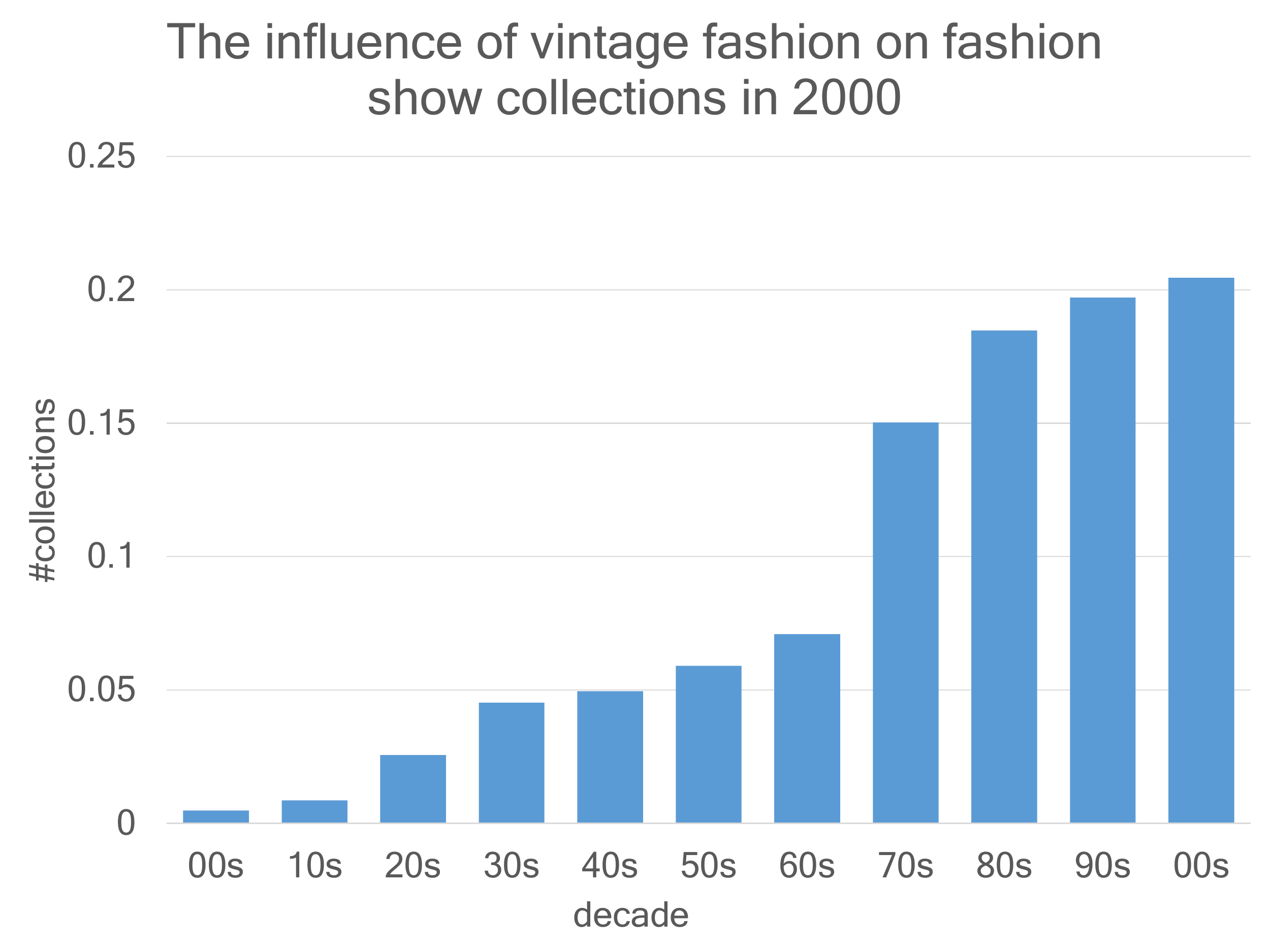}	
			\caption{2000}
			\label{fig:influence_01}
	\end{subfigure}
	\begin{subfigure}[b]{0.32\textwidth}
		\centering
			\includegraphics[width=\textwidth]{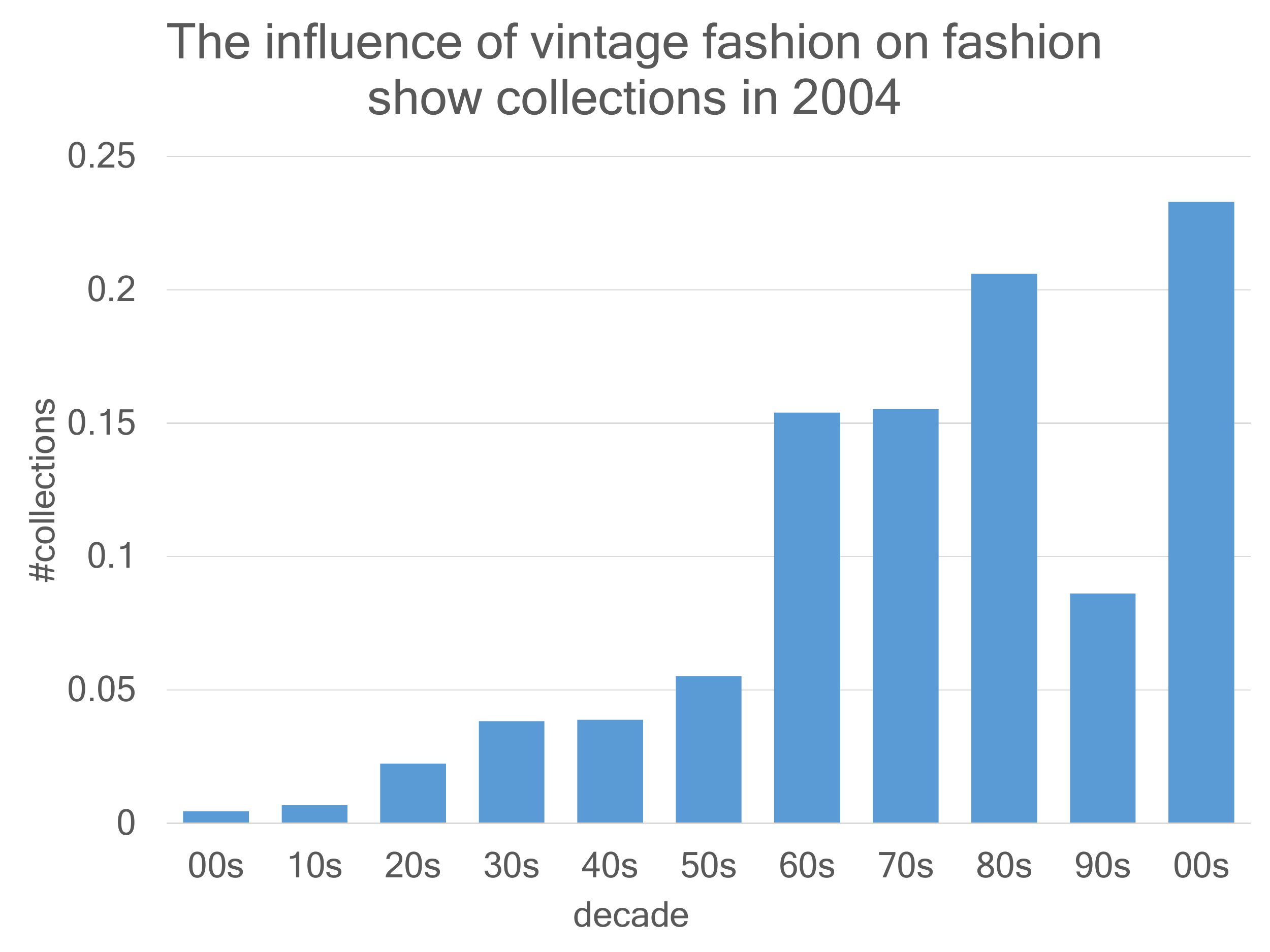}	
			\caption{2004}
			\label{fig:influence_05}
	\end{subfigure}
	\begin{subfigure}[b]{0.32\textwidth}
		\centering
			\includegraphics[width=\textwidth]{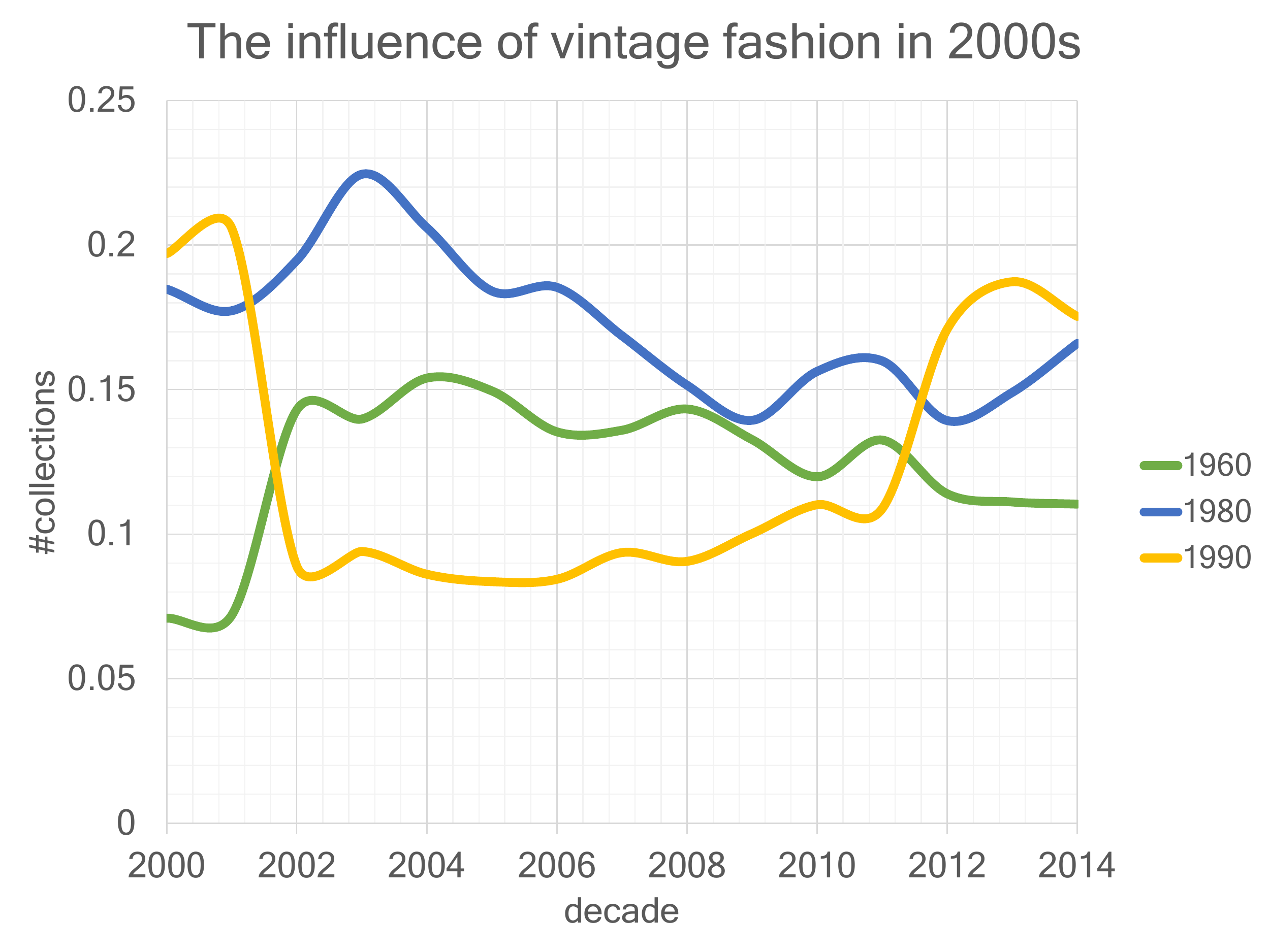}	
			\caption{2000-14}
			\label{fig:influence_16}
	\end{subfigure}
	\caption{The influence of vintage fashion (1900s-2000s) in fashion show collections from (a) 2000 and (b) 2004. 
Figure (c) shows the influence of 1960s, 1980s and 1990s fashion trhough out 2000s - 2010s.}
\label{fig:app2}
\end{figure*}

\vspace{-.2cm}
\section{Applications}
\label{sec:app}

Denim miniskirts, ripped distressed jeans, denim jackets, tracksuits, trench coats, puffy jackets, preppy polo shirts with popped collars, neon colors, gladiator shoes, the list can go on and on. People look back to the past and say ``That's ridiculous! What were they wearing?''. Yet somehow the fashion world recycles these trends, makes a few tweaks, and voila! they keep coming back. It's nearly impossible to think of something that has absolutely never been done before. Designers have to look for inspiration from somewhere, and what better place to be inspired by  than the past?

Therefore, we demonstrate the use of our learned models for predicting the influence of fashion from past decades on the fashion collections. In particular, we evaluate our models on a Runway dataset~\cite{runway2realwayWACV15} containing 300k images from fashion shows over the years of 2000 to 2015. To estimate the influence of the vintage fashion on fashion collections, we first apply our fine-tuned model to estimate when the outfits were made. Then, we define the inspiration date of the collection as the decade with the highest probability among images from that collection. To evaluate our approach, we collect human judgments on the same task using Amazon Mechanical Turk. For each assignment, 5 works are shown 5 fashion show images per collection and ask to identify the decade that inspired these images. We randomly pick up to 200 collections per predicted decade and remove collections with low human agreement (less than 3 of 5 agree). This leaves us with 300 collections for evaluation. 
On these collections, our model has $58.33\%$ agreement with MAE of $8.6$ years compared to human judgments. Some examples of our temporal prediction are shown in Figure~\ref{fig:app}. 

Finally, we also observe temporal influence trends on fashion show collections over time. To do so, we explore the classification confidence of collections from the same year as shown in Figure \ref{fig:influence_01} - \ref{fig:influence_05}. 
If we look at the classification confidence of a particular vintage decade across years, we spot some interesting trends. For example, from Figure \ref{fig:influence_16}, we can see that 1990s fashion had a strong influence during both early 2000s and early 2010s while during the mid-2000s a revival of 1960s and 1980s fashion occurred~\cite{2000sFashionWiki}.

\vspace{-.2cm}
\section{Conclusions}
\vspace{-.1cm}
In this work, we first explore CNN approaches to automatically estimate when objects were made, evaluated on an existing dataset of cars and two new datasets of vintage clothing photographs. Then, we provide several analyses of what the networks have learned, including exploring the temporal sensitivity of nodes, as well as examining node activations and comparison to discriminative parts learned by the data mining approach~\cite{lee_styleAware}. Finally, we propose an application for temporal estimation task of clothing in the real world scenario.
{\small
\bibliographystyle{ieee}
\bibliography{draft_01}
}
\end{document}